\documentclass[conference]{IEEEtran}
\IEEEoverridecommandlockouts
\usepackage{cite}
\usepackage{amsmath,amssymb,amsfonts}
\usepackage{graphicx}
\usepackage{textcomp}
\usepackage{float}
\usepackage{xcolor}
\usepackage{subfigure}
\usepackage{caption}
\usepackage{multicol}
\usepackage{amsmath}
\usepackage{enumitem}
\usepackage{comment}
\usepackage{multirow}
\usepackage[ruled,vlined]{algorithm2e}

\def\BibTeX{{\rm B\kern-.05em{\sc i\kern-.025em b}\kern-.08em
    T\kern-.1667em\lower.7ex\hbox{E}\kern-.125emX}}
\begin{document}

\title{Vanishing Bias Heuristic-guided Reinforcement Learning Algorithm}
\author{
    \IEEEauthorblockN{\textbf{Qinru Li}\IEEEauthorrefmark{2}, \textbf{Hao Xiang}\IEEEauthorrefmark{2}}
    \IEEEauthorblockA{\IEEEauthorrefmark{2}University of California, San Diego}
}
\maketitle
\begin{abstract}
     Reinforcement Learning has achieved tremendous success in the many Atari games. In this paper we explored with the lunar lander environment and implemented classical methods including Q-Learning, SARSA, MC as well as tiling coding. We also implemented Neural Network based methods including DQN, Double DQN, Clipped DQN. On top of these, we proposed a new algorithm called Heuristic RL which utilizes heuristic to guide the early stage training while alleviating the introduced human bias. Our experiments showed promising results for our proposed methods in the lunar lander environment. 
\end{abstract}

\section{Introduction}

The idea of learning through interaction is probably one of the most natural ways to learn. Though touching, seeing, speaking, and acting, infants can understand their surrounding environment without any explicit instruction from parents. As we can learn from experience, one may wonder that how could we extend this capability to machine. The study of this topic is called reinforcement learning (RL). In the setting of the reinforcement learning, the learner is called agent. The agent is able to sense the environment to some extend and take actions. The goal of the agent is to figure out a set of actions based on its observation of the environment and maximize its reward defined in the environment. Reinforcement Learning has achieved many success in domain like multi-robot navigation, autonomous driving and games. 

In this paper, we are going to study reinforcement learning algorithms in a game environment: Lunar Lander. The goal of the game is to control the rocket of the lander to land on the moon safely. The environment is formulated into a Markov decision process (MDP). We implemented three classic reinforcement learning algorithms: Monte Carlo control algorithm; an variant of temporal difference (TD) algorithm - SARSA; and Q-learning algorithm. We also implemented several deep learning based reinforcement learning algorithm: DQN, Double DQN, Clipped Q-Learning as well as their 3 heuristic variants. 

Our \textbf{main contributions} for this paper can be summarized as follows. \\
\begin{enumerate}
    \item For classical RL algorithm, we discretize the environment and augment the feature representation with \textbf{tile coding}. 
    \item We implemented three classical RL algorithms: \textbf{Q-Learning, SARSA, Monte Carlo Control}. 
    \item We proposed a new algorithm called \textbf{heuristic RL} for the early stage of training, which can efficiently prune the search space and alleviate the human bias introduced by heuristic. And our experiment demonstrate promising results. 
    \item We implemented deep learning based RL algorithms \textbf{DQN, Double DQN, Clipped Q-Learning} and their heuristic variants based on Vanilla DQN implementation.
\end{enumerate}

\section{Related work}
 Neural network (NN) has demonstrated its success in many domains like Natural Language Processing \cite{houlsby2019parameter}, few-shot learning \cite{garcia2017few}, face detection \cite{rowley1998neural} etc. Classical Q-learning uses matrix to keep track of the intermediate values. However when the state space is very large or even continuous, Q-table can consume all the memory, leading to the difficulty of training. To overcome this problem, we can use value function approximation. Classical methods usually relies on well-designed hand-crafted features. Even though they are easy to train when the function is simple like linear, they have limited capacity for complex nonlinear function. On the contrary, deep learning can utilize highly non-convex functions to approximate the value function. And according to universal approximation theory, even a feed-forward neural network can approximate arbitrary real-value functions, making it a promising candidate for q-value approximation. \cite{mnih2013playing} proposed to utilize Deep Q-Networks (DQN) to estimate the value function. To overcome its over estimation problem, \cite{hasselt2010double} utilized double q learning which separate the action value evaluation and action selection. \cite{fujimoto2018addressing} utilized a clipped Double Q-learning to alleviate the overestimation. 
 
 Data-driven Reinforcement Learning learns in a zigzag way. Especially during the early stage of training, the robot may find it difficult to find one single positive sample among the huge search space. For some game, the agent can receive rewards only when the game is over. The exploration and exploitation technique can become useful when the agent becomes on the track of learning something. For the early stage, the robot may just get stuck of learning shown in Figure \ref{fig:2}. All those difficulties bring up \textbf{two questions}:
 \begin{enumerate}
     \item  How can we efficiently utilize prior knowledge to help the robot learn positive samples during the early stage of training?
     \item How can we utilize the prior knowledge without introducing too much human bias?
 \end{enumerate}
  \cite{keselman2018reinforcement} proposed to utilize a deep neural network to learn the heuristic. But we argue that using neural network to learn heuristic is more like adding features than using real prior knowledge. The most similar work is \cite{xie2020heuristic}, they proposed to utilize a customized heuristic reward function for UAV. However they fail to address the second question. Using heuristic for all the stage of training would introduce strong human bias. Thus the performance of the algorithm depends largely on the quality of the hand-crafted heuristic functions. Differently, we propose using a heuristic-guided method only for the early stage of training. During the early stage, the robot can quickly find a feasible solution based on heuristic, which could prune the search space and speed up the training. Once the algorithm has gained sufficient positive samples, we use a decaying technique to reduce the bias and heuristic. In other words, after early stage training, the algorithm learn in a data-driven fashion and the bias gradually vanishes. 

\section{Problem Formulation}
\textbf{Reinforcement Learning Formulation}
Under MDP assumptions with unknown motion model  $p_{f}\left(\cdot | \mathbf{x}, \mathbf{u}\right)$ and reward function $l(\mathbf{x}, \mathbf{u})$,  we want to optimize the policy $\pi^*$ s.t.
\begin{equation}
\pi^* =\arg \max _{\pi_{\tau: T-1}} V_{\tau}^{\pi}\left(\mathbf{x}_{\tau}\right)
\end{equation}
\begin{equation}
 V_{\tau}^{\pi}\left(\mathbf{x}_{\tau}\right) :=\mathbb{E}_{\mathbf{x}_{\tau+1: T}}\left[\sum_{t=\tau}^{T-1} \gamma^{t-\tau} \ell\left(\mathbf{x}_{t}, \pi_{t}\left(\mathbf{x}_{t}\right)\right) | \mathbf{x}_{\tau}\right]
\label{eqn:obj_func_generic}
\end{equation}
where 
\begin{align*}
& t=\tau, \ldots, T-1 \\ 
& \mathbf{x}_{t+1} \sim p_{f}\left(\cdot | \mathbf{x}_{t}, \pi_{t}\left(\mathbf{x}_{t}\right)\right) \\
& \mathbf{x}_{t} \in \mathcal{X} \\
& \pi_{t}\left(\mathbf{x}_{t}\right) \in \mathcal{U}\left(\mathbf{x}_{t}\right)
\end{align*}
\begin{enumerate}
	\item $\mathcal{X}\subseteq R^8$ represent the state space. For $x\in \mathcal{X}$, $x_1,x_2$ is the 2D coordinate of the robot. $x_3,x_4$ is the horizontal and vertical speeds. $x_5,x_6$ is the angle and angular velocity. $x_7,x_8$ represents whether the left/right leg of the robot has contact with the ground. $\mathcal{U}=\{0,1,2,3\}$ be the discrete set of control space including. The details of control space and state space are discussed in the game environment section. And the state space can be continuous for neural network based algorithm and discretized for classical algorithms we experimented for this project. 
	\item Denote the planning horizon as $T$. Depending the methods and algorithms we can formulate it as infinite horizon or finite horizon.
	\item Instead of using cost to represent the value of the state and action, we use reward here. The stage reward is denoted as $l(\mathbf{x}_t.\mathbf{u}_t)$ for given $\mathbf{x}_{t} \in \mathcal{X}$ and $\mathbf{u}_{t} \in \mathcal{U}$ at time $t$. If it is finite horizon, then the terminal reward is 0.
	\item Denote the discount factor as $\gamma$ and $\gamma \in(0,1)$.
\end{enumerate}

Assume that we are always starting from $t=0$, then the equation can be simplified to 
\begin{equation}
V^{*}(\mathbf{x})=\max _{\pi} V^{\pi}(\mathbf{x}):=\mathbb{E}\left[\sum_{t=0}^{\infty} \gamma^{t} \ell\left(\mathbf{x}_{t}, \pi_{t}\left(\mathbf{x}_{t}\right)\right) | \mathbf{x}_{0}=\mathbf{x}\right]
\label{eqn:obj_func_simplified}
\end{equation}

\section{Technical Approach}





In this section, we will describe the game environment and our algorithms to control the agent in the environment. We will present three classical reinforcement learning algorithms: Q-learning, SARSA, and Monte Carlo control algorithm and three variants of the deep learning based algorithm based on vailla DQN implementation\footnotemark. We also proposed a new algorithm called Heuristic DQN, which could efficiently utilize heuristics to guide the learning. From the best of our knowledge, we are \textbf{the first} to propose utilizing vanishing bias non-learnable heuristic functions for DQN.
\footnotetext{https://github.com/ranjitation/DQN-for-LunarLander}

\subsection{Game Environment}
The game environment we are using is the Lunar Lander V2 developed by OpenAI\footnotemark. In the game, there are only one agent: the lander. Our goal is to control the lander so that it can safely land on a landing pad in the middle of two flags. The landing pad is always placed in the center of the environment. The initial velocity of the lander is random for every game play. Therefore, the lander cannot cheat to win the game by doing nothing. The game would be over if the lander crashes or comes to rest. 
\footnotetext{https://gym.openai.com/envs/LunarLander-v2/}

The state of the agent is represented by eight variables $x=\{p_x, p_y, v_x, v_y, \theta, v_\theta, lg_1, lg_2\}$. $p_x, p_y\in \mathbb{R}$ indicate the x, y position of the agent, x is the horizontal position. y is the vertical position. The origin of the xy coordinate is at the center of the landing pad. $v_x, v_y \in \mathbb{R}$ are the velocity of the agent in the x, y directions. $\theta $ is the orientation of the agent represented in radiant. $v_\theta \in \mathbb{R}$ is the angular velocity of the agent. $lg_1, lg_2$ are boolean variables, where if the left leg of the lander has contact with the ground, then $lg_1$ is true. Otherwise $lg_1$ is False. Similar relation applies to right leg and $lg_2$. In the game environment, the feasible range of $p_x$, $p_y$ is $[-1, 1]$. If the agent fly outside this region, the game will terminate. The initial random speed of the agent is guaranteed to be finite. We can assume that $v_x, v_y, v_\theta \in [-10^4, 10^4]$. 

The lander has four actions to control its position: fire left orientation engine, fire right orientation engine, fire main engine (which pointing down), and do nothing. The fire engine is either in full throttle or is turned off and the lander is assumed to have infinite fuel. We encode the action into 4 discrete integers. 
\begin{align*}
u =
    \begin{cases}
    0, & \text{Do nothing} \\
    1, & \text{Fire left engine} \\
    2, & \text{Fire down engine} \\
    3, & \text{Fire right engine} \\
    \end{cases}
\end{align*}

The motion module of the agent is handled by the OpenAI for which we are assume unknown. The reward of an action depend on the state of the lander. If the lander crashes, the reward is -100, but if the lander manages to land on the surface, the reward is +100. Each leg ground contact is +10. Firing the main engine is -0.3 to encourage lander to use engine as less as possible. Firing other engines incurs no reward. A successfully landing in the landing pad (i.e. two leg of the lander touch the landing pad) has an additional +200 reward. 

For classical methods, we discretize the state into $k$ layers of 9-D grid by tile coding. The 9 dimension is the summation of the 8 dimension state space and 1 dimension of the action space. The shape of each grid is the same. We discretize the state $x$ for each grid with different offset. Denote the resolution of the state as $r_x = \{r_{p_x}, r_{p_y}, r_{v_x}, r_{v_y}, r_{\theta},r_{v_\theta},r_{lg_1}, r_{lg_2}\}$, where each element in $r_x$ is the resolution of the corresponding variable in $x$. Because $lg_1, lg_2$ are boolean variable, we do not need to define their resolution, so $r_{lg_1}=r_{lg_2}=0$ and they do not participate in the division (to avoid the divided-by-zero case). The coordinate $c_x^i$ of $x$ in grid $i$ is calculated as $c_x^i=x-(i - 1)\cdot r_x / k$. Note that the index of grid starts from 1. The index $I_x^i$ of $c_x^i$ in the grid is $I_x^i=round(c_x^i / r_x)$ where $round(\cdot)$ is a rounding function that maps any elements in the given vector to its nearest integer. A tile coding with only one layer of such grid is often called bucketization. 

We store the Q value of each state and action pair $(x, u)$ into the grids. Our access pattern to the Q value function is similar to \cite{Whiteson2007AdaptiveTC}. Each layer has an associated weight $w_i$, $\sum_{i=1}^{k} w_i=1$, so the final Q value of $(x, u)$ would be the weight sum of the value of each grid. We select the uniform distribution for $w_i$, that is $w_1=w_2=...=w_k=1/k$. The update rule follows the similar pattern. These two algorithms are presented in Algorithm \ref{alg:get from tile} and \ref{alg:set from tile}. 

\begin{algorithm}
    \caption{GetFromTile$(x, u)$}
    \label{alg:get from tile}
    \SetAlgoLined
    $Q\leftarrow 0$\;
    \For{Each grid $G_i$} {
        $c_x^i=x-(i - 1)\cdot r_x / k$\;
        $I_x^i=round(c_x^i / r_x)$\;
        $Q\leftarrow Q + w_i\cdot G_i[I_x^i, u]$
    }
    \Return {$Q$}
\end{algorithm}

\begin{algorithm}
    \caption{UpdateTile($\Delta Q(x, u)$)}
    \label{alg:set from tile}
    \SetAlgoLined
    \For{Each grid $G_i$} {
        $c_x^i=x-(i - 1)\cdot r_x / k$\;
        $I_x^i=round(c_x^i / r_x)$\;
        $G_i[I_x^i, u] = G_i[I_x^i, u] + w_i \cdot \Delta Q(x, u)$\;
    }
\end{algorithm}


\subsection{Q Learning}

The Q learning algorithm is presented in Algorithm \ref{alg:q learning}. The initial Q value of each $(x, u)$ is 0. We use $\epsilon$ - greedy algorithm to generate the action $a_t$. That is given an $\epsilon_t \in[0, 1]$ at time $t$, the policy of state $x$ is generated by 

\begin{equation}
\pi_{t}(\mathbf{u} | \mathbf{x}):=\left\{\begin{array}{ll}
1-\epsilon_{t}+\frac{\epsilon_{t}}{|\mathcal{U}(\mathbf{x})|} & \text { if } \mathbf{u}=\mathbf{u}^{*} \\
\frac{\epsilon_{t}}{|\mathcal{U}(\mathbf{x})|} & \text { if } \mathbf{u} \neq \mathbf{u}^{*}
\end{array}\right.
\end{equation}
where $\mathbf{u}^{*}$ is the control with the largest reward for state $x$. By doing so, the algorithm can both exploit its learned value function and explore a new state-action, which will make our algorithm generalize better. Note that the initial state of the lunar lander is also not constant, which will provide a similar benefit to the exploration and exploitation. Therefore, here we do not need a high $\epsilon_{t}$ value. 

$\alpha_t$ is the learning rate of the algorithm at time $t$. In order for the Q learning algorithm to converge, $\alpha_t$ must satisfy the Robbins-Monro criterion. That is 
\begin{equation}
\sum_{t=1}^{\infty} \alpha_{t}=\infty \quad \sum_{t=1}^{\infty} \alpha_{t}^{2}<\infty
\end{equation}
We use $\alpha_t = c/(c+t)$ as our learning rate decay function. Here $c\in R, c>0$ is a positive constant. It can be shown that this algorithm satisfies the Robbins-Monro criterion. 

\begin{algorithm}
    \caption{Q Learning}
    \label{alg:q learning}
    \SetAlgoLined
    Initialize the Q value of all the state to zero\;
    \For{$i\leftarrow 1$ to max iterations}{
        \While{game is not done and not exceed max steps}{
            Sample $a_t\sim \pi^\epsilon(x_t)$\;
            Execute $a_t$ in env and observe $x_{t+1},r_t,o_t$\;
            $Q(x_t, a_t)$ = GetFromTile($x_t, a_t$)\;
            $\Delta Q(x_t, a_t)=\alpha_t (r_t+\gamma \max_{a'}Q(x_{t+1},a') - Q(x_t, a_t))$\;
            UpdateTile($\Delta Q(x_t, a_t)$)\;
        }
    }
\end{algorithm}

\subsection{SARSA}
Unlike the Q learning algorithm, where the update of the $Q(x_{t}, u_{t})$ depend on the estimated best action of the next state $x_{t+1}$, SARSA uses the temporal difference updates after every $x_{t}, u_{t}, r_t, x_{t+1}, u_{t+1}$ transition to update the $Q(x_{t}, u_{t})$. We use the same $\epsilon$ - greedy algorithm to balance the exploration and exploitation, and the same earning rate decay function to insure that the algorithm will converge. The algorithm is presented in Algorithm \ref{alg:sarsa}.

\begin{algorithm}
    \caption{SARSA}
    \label{alg:sarsa}
    \SetAlgoLined
    Initialize the Q value of all the state to zero\;
    $S\leftarrow \emptyset$\;
    \For{$i\leftarrow 1$ to max iterations}{
        \While{game is not done and not exceed max steps}{
            Sample $a_t\sim \pi^\epsilon(x_t)$\;
            Execute $a_t$ in env and observe $x_{t+1},r_t,o_t$\;
            
            \If {$S$ not is empty} {
                $x_{t-1}, a_{t-1}, r_{t-1}\leftarrow S$\;
                $Q(x_{t-1}, a_{t-1})$ = GetFromTile($x_{t-1}, a_{t-1}$)\;
                $\Delta Q(x_{t-1}, a_{t-1})=\alpha_t (r_{t-1}+\gamma Q(x_{t},a_{t}) - Q(x_{t-1}, a_{t-1}))$\;
                UpdateTile($\Delta Q(x_{t-1}, a_{t-1})$)\;
            }
            $S\leftarrow \{x_{t}, a_{t}, r_{t}\}$\;
        }
    }
\end{algorithm}

\subsection{Monte Carlo}

Monte Carlo (MC) algorithm is another method to estimate the value function of state and action when the model is unknown. It uses the empirical mean of long-term costs obtained from different episodes to approximate the Q value $Q(x, u)$. There are two kinds of Monte Carlo policy evaluation algorithm. One is the first-visit MC policy evaluation and another one is every vist MC policy evaluation. For the former one, it find the first time step when $x, u$ appear and calculate their expected long term cost, while for the latter one, the long-term costs are accumulated following every time step $t$ when $x, u$ appear. The two algorithms are shown in Algorithm \ref{alg:monte carlo first visit} and \ref{alg:monte carlo every visit}. 

Similar to the Q learning algorithm and the SARSA algorithm, we use the same learning rate decay function and the $\epsilon$ - greedy algorithm for the MC algorithm. 

\begin{algorithm}
    \caption{Monte Carlo - First Visit}
    \label{alg:monte carlo first visit}
    \SetAlgoLined
    Initialize the Q value of all the state to zero\;
    \For{$i\leftarrow 1$ to max iterations}{
        $S\leftarrow \emptyset$\;
        \While{game is not done and not exceed max steps}{
            Sample $a_t\sim \pi^\epsilon(x_t)$\;
            Execute $a_t$ in env and observe $x_{t+1},r_t,o_t$\;
            $S \leftarrow S \cup {(x_t, a_t, r_t)}$\;
        }
        
        $L \leftarrow 0$\;
        \For {$t$ from $T$ to $1$} {
            $x_{t}, a_{t}, r_{t}\leftarrow S(t)$\;
            $L \leftarrow \gamma L + r_t$\;
            \If {This is the first occurrence of $(x_t, u_t)$} {
                $Q(x_t, a_t)$ = GetFromTile($x_t, a_t$)\;
                $\Delta Q(x_t, a_t)=\alpha_t (L - Q(x_t, a_t))$\;
                UpdateTile($\Delta Q(x_t, a_t)$)\;
            }
        }
    }
\end{algorithm}

\begin{algorithm}
    \caption{Monte Carlo - Every Visit}
    \label{alg:monte carlo every visit}
    \SetAlgoLined
    Initialize the Q value of all the state to zero\;
    \For{$i\leftarrow 1$ to max iterations}{
        $S\leftarrow \emptyset$\;
        \While{game is not done and not exceed max steps}{
            Sample $a_t\sim \pi^\epsilon(x_t)$\;
            Execute $a_t$ in env and observe $x_{t+1},r_t,o_t$\;
            $S \leftarrow S \cup {(x_t, a_t, r_t)}$\;
        }
        
        $L \leftarrow 0$\;
        $M$ maps state-action pair to a list of rewards\;
        \For {$t$ from $T$ to $1$} {
            $x_{t}, a_{t}, r_{t}\leftarrow S(t)$\;
            $L \leftarrow \gamma L + r_t$\;
            $M(x_t, u_t) \leftarrow M(x_t, u_t) + L$\;
        }
        \For {Every $(x_t, u_t)$ in $M$} {
            $L_{avg} = avg(M(x_t, u_t))$\;
            $Q(x_t, a_t)$ = GetFromTile($x_t, a_t$)\;
            $\Delta Q(x_t, a_t)=\alpha_t (L_{avg} - Q(x_t, a_t))$\;
            UpdateTile($\Delta Q(x_t, a_t)$)\;
        }
    }
\end{algorithm}

\subsection{Deep Q-Learning}
The algorithm for DQN is shown Algorithm\ref{alg:1}. Notice that except that DQN utilized neural networks to estimate the Q-value, there are still some remarkable difference between DQN and Classical Q-learning. For DQN, we use experience reply technique to reduce the correlations between data samples. And DQN utilized  two different networks for evaluation and selection. And the update steps for target network parameters is soft update, i.e.
\begin{equation}
    \theta' \rightarrow \tau \theta +(1-\tau)\theta'
\end{equation}
In this way, the network can better capture the temporal information from previous training and only gradually change its parameters. The sampling policy $\pi^\epsilon$ is $\epsilon$-greedy policy. 
\begin{algorithm}
\SetAlgoLined
Initialize local network $Q_{\theta}$ and target network $Q_{\theta'}$\;
Initialize experience reply buffer $D$\;
\For{$i\leftarrow 1$ to max iterations}{
    \While{game is not done and not exceed max steps}{
    Sample $a_t\sim \pi^\epsilon(x_t)$\;
    Execute $a_t$ for env and observe $x_{t+1},r_t,o_t$\;
    Store $(x_t,a_t,r_t,x_{t+1},o_t)$ to $D$\;
    Periodically do the following updates\;
    Sample $(x_t,a_t,r_t,x_{t+1},o_t)\sim D$\;
    $\hat Q(x_t,a_t)=r_t+\gamma \max_{a'}Q(x_{t+1},a';\theta')*(1-o_t)$\;
    Do SGD for $\ell=\left(\hat Q(x_t,a_t)-Q(x_t,a_t;\theta) \right)^2$\;
    Update target network parameters $\theta'$\;
    }
}
 \caption{DQN}
 \label{alg:1}
\end{algorithm}

\subsection{Double DQN}
The core idea for Double DQN is to decouple the action evaluation and action selection. Double DQN uses target network to select the action and use another local network to evaluate the Q-value. In such a way, we can avoid overestimation problems of action\cite{hasselt2010double,van2016deep}. And it would help the network to better learn unbiased values. The algorithm for Double DQN is shown in Algorithm\ref{alg:2}\\
\begin{algorithm}
\SetAlgoLined
Initialize local network $Q_{\theta}$ and target network $Q_{\theta'}$\;
Initialize experience reply buffer $D$\;
\For{$i\leftarrow 1$ to max iterations}{
    \While{game is not done and not exceed max steps}{
    Sample $a_t\sim \pi^\epsilon(x_t)$\;
    Execute $a_t$ for env and observe $x_{t+1},r_t,o_t$\;
    Store $(x_t,a_t,r_t,x_{t+1},o_t)$ to $D$\;
    Periodically do the following updates\;
    Sample $(x_t,a_t,r_t,x_{t+1},o_t)\sim D$\;
    $a^*=\arg \max_{a}Q(x_{t+1},a;\theta')$
    $\hat Q(x_t,a_t)=r_t+\gamma Q(x_t,a^*;\theta)*(1-o_t)$\;
    Do SGD for $\ell=\left(\hat Q(x_t,a_t)-Q(x_t,a_t;\theta) \right)^2$\;
    Update target network parameters $\theta'$\;
    }
}
 \caption{Double DQN}
 \label{alg:2}
\end{algorithm}
\subsection{Clipped Double Q-Learning}
Another variant of Double DQN we implemented is a variant of Clipped Double Q-learning\cite{fujimoto2018addressing}. We call it Clipped DQN for short. The algorithm is shown in Algorithm\ref{alg:3}. When calculating the target q-value, like double q learning, we separate action selection and action value evaluation. Unlike double Q learning, Clipped DQN takes the minimum of the q-values estimated by two networks. In such a way, the algorithm can alleviate the overestimation problems and provide more stable estimations\cite{fujimoto2018addressing}. 
\begin{algorithm}
\SetAlgoLined
Initialize local network $Q_{\theta}$ and target network $Q_{\theta'}$\;
Initialize experience reply buffer $D$\;
\For{$i\leftarrow 1$ to max iterations}{
    \While{game is not done and not exceed max steps}{
    Sample $a_t\sim \pi^\epsilon(x_t)$\;
    Execute $a_t$ for env and observe $x_{t+1},r_t,o_t$\;
    Store $(x_t,a_t,r_t,x_{t+1},o_t)$ to $D$\;
    Periodically do the following updates\;
    Sample $(x_t,a_t,r_t,x_{t+1},o_t)\sim D$\;
    $\hat Q(x_t,a_t)=r_t+\min_{w\in\{\theta',\theta\}}\gamma \max_{a'}Q(x_{t+1},a';w)*(1-o_t)$\;
    Do SGD for $\ell=\left(\hat Q(x_t,a_t)-Q(x_t,a_t;\theta) \right)^2$\;
    Update target network parameters $\theta'$\;
    }
}
 \caption{Clipped Double Q Learning}
 \label{alg:3}
\end{algorithm}

\subsection{Heuristic-guided algorithms}
After experimenting with all the aforementioned methods, we find that those algorithms learn very slowly during the early stage of training. This also agree with our human intuition that when the robot know nothing about the task, the robot can only randomly guess the best action to take, thus during early stage, robot can only learn positive rewards when it lands on the moon successfuly, which is a very rare case for the most of the time. Inspired by $A^*$ algorithms, we introduced heuristic functions to help the robot better navigate the environment during the early stage of training. 

The robot can only receive positive rewards when it interacts with the ground and landing pads. But during early stage of training, the robot can hardly find the path to the landing pads and sometimes the robot may just float in the air. Thus we need some heuristics to guide the robot. We defined the following heuristic functions.
\begin{align}
    h(s_t,s_{t+})&=\left\{\begin{array}{cc}
         k_1*\mathbf{\phi}(s_t,s_{t+1})&s_{t+1}\in \mathcal{B}_{t^*}^{\epsilon_1}  \\
          k_2*\mathbf{\phi}(s_t,s_{t+1})&s_{t+1}\not\in \mathcal{B}_{t^*}^{\epsilon_1}
    \end{array} \right.\\
    \mathbf{\phi}(a,b)&=\alpha \phi_1\left(\begin{bmatrix}a_x\\a_y \end{bmatrix}, \begin{bmatrix}b_x\\b_y \end{bmatrix}\right)+\beta  \phi_2\left(\begin{bmatrix}a_{\theta_x}\\a_{\theta_y} \end{bmatrix}, \begin{bmatrix}b_{\theta_x}\\b_{ \theta_y} \end{bmatrix}\right)
\end{align}
where $0<k_1\ll k_2$, $0\le \beta \ll \alpha$ and $\mathcal{B}_{t}^{\epsilon_1}$ is the set of points that are close to the landing pad.
\begin{align}
    \phi_1\left(\begin{bmatrix}a_x\\a_y \end{bmatrix}, \begin{bmatrix}b_x\\b_y \end{bmatrix}\right)&=b_x^2+b_y^2
\end{align}
And $\phi_2\left(\begin{bmatrix}a_{\theta_x}\\a_{\theta_y} \end{bmatrix}, \begin{bmatrix}b_{\theta_x}\\b_{ \theta_y} \end{bmatrix}\right)$ is the relative angle change with respect to the vertical axis, which measures the orientation change. For analysis simplicity, here we assume the vertical axis is perpendicular to the landing pad. Generally, we need to transform the coordinates so that the y-axis is perpendicular to the landing pad in order to calculate the error.

Heuristic function $h$ measures the advantage of transforming from state $s_t$ to $s_{t+1}$. $\phi_1$ measures the distance from $s_{t+1}$ to the goal $t^*$. Intuitively, if the new state is closer to the goal, then we have more advantage. $\phi_2$ measures the angle changes. If the angle is far away from the angle of the terminal state, then we have less advantage when taking action $a_t$. The heuristic function is utilized in the step of calculating target q-value. 
\begin{align}
    \hat Q(x_t,a_t)=r_t-\alpha_t h(x_t,x_{t+1})+\gamma \max_{a'}Q(x_{t+1},a';\theta')*(1-o_t)
\end{align}
For the early stage training, the rewards for most action value pairs are similar and the long term estimation $Q(x_{t+1},a';\theta')$ is not accurate. Thus by introducing the heuristic function, we can take advantage of heuristic distinguish two actions which have similar stage costs. The heuristic function can guide the robot to move towards the goal $s_0=(0,0)$ and adjust its angle so that the robot is parallel to the landing pad.\\
\textbf{Vanishing Bias}\\
Heuristic functions will introduce human bias and usually leads to local minimum. If the robot trust heuristic too much, then the algorithm does not fully utilize the great power of function estimation of neural networks. Thus as we have discussed, the heuristics are only used for the early stage training. Once the network has learned useful information, we should again trust the estimated long term rewards. Thus we introduced decaying factor $\alpha_t$. Periodically, we set $\alpha_{t+1}=p* \alpha_t$ ($0<p<1$) to lessen the importance of heuristic during later training. In such a  way, the human bias introduced by heuristic will gradually vanish. The network will trust more on its estimation and learn from the data sampled from the experience reply buffer. The main point for our proposed approach is to help the robot quickly find a feasible way to success. And let the network learn from those positive samples. Once the early stage guidance is done, we should again trust the network estimation and follow the general data-driven approach. In the experiments part, we will show the promising results for our proposed methods. 

To generalize this approach to all the aforementioned methods, we can simply replace $r_t$ with $r_t-\alpha_th(x_t,x_{t+1})$ when calculating $\hat Q(x_t,a_t)$.

\section{Experiments}




\subsection{Metrics}
Denote $N$ as the total number of trails. And denote $R_i$ as the cumulative score for trial $i$. Denote the final reward for the agent is $r_i,r_i\le100$ for the trial $i$. Denote the number of times the agent fires as $u_i$. For notation simplicity, we abuse the notation for this subsection. All the other part of the paper follows the notation we defined before.
\begin{enumerate}
    \item \textbf{Average scores }Average score is calculated by taking the mean of scores for each trial. It measures how good the agent plays on average.
    \begin{align}
        \text{Average scores}=\frac{\sum_{i=1}^n R_i}{N}
    \end{align}
    \item \textbf{Probability of success(PoS) }Following lunar lander environment's requirement, we define that a agent succeeds if its final score is greater than 200 points. Thus Probability of success can be calculated by 
    \begin{align}
        PoS&=\frac{N_{success}}{N}
    \end{align}
    where $N_{success}$ is the total number times a robot succeeds.
    \item \textbf{Average fuel consumption }It measures how many times the robot fires, which represents how many fuels a robot would need for each trail.
    \begin{align}
        \text{Average fuel consumption}=\frac{\sum_{i=1}^N u_i}{N}
    \end{align}
    \item \textbf{Average terminal score} It measures the average terminal score and can be calculated as follows.
    \begin{align}
        \text{Average terminal score}=\frac{\sum_{i=1}^n r_i}{N}
    \end{align}
\end{enumerate}

\subsection{Neural Network Setup}
We use fully connected layer for the Deep Q-Network with hidden units 256, 128, 64, 4. The activation function is ReLU. We experimented with batch size of 64, 128 and 1024. And we find that when using batch size 64, we have better results. The learning rate ranges from $10^{-4}$ to $10^{-3}$. We use Adam optimizer. The experience reply buffer size is $10^4$. For updating network parameter step, we set $\tau=0.001$. For the heuristic algorithms, we set $k_1=1,k_2=0.1, \alpha=1, \beta=0.1$. Also the decay factor $\alpha_0=100$ and we decay $\alpha_t$ every 10 iterations. The training step follows the Algorithm\ref{alg:1},\ref{alg:2},\ref{alg:3}. Also we adopted a similar approach as early stop. When the average scores of the game is greater than a threshold for a certain amount of episodes, we stop the training. For the testing part, we generate another set of environments and let the agents play the games and calculate corresponding metrics. We do 100 experiments for each agent during the testing phase.

\begin{table*}[t]
    \centering
    \begin{tabular}{|c|c|c|c|c|}
        \hline
        Algorithm&Average fuel consumption&Average terminal score&Average score&Probability of success\\
        \hline
        \hline
        Q Learning&271.90&-11.98&65.09&0.26 \\
        \hline
        SARSA&331.02&-19.16&25.41&0.10 \\
        \hline
        Monte Carlo &0.72&-100.00&-133.92&0.00 \\
        \hline
        DQN&354.8&82.02&198.84&0.63 \\
        \hline
        DQN+Heuristic&201.34&\textbf{99.0}&\textbf{258.23}&\textbf{0.99} \\
        \hline
        Double DQN&\textbf{141.6}&11.99&104.82&0.5 \\
        \hline
        Double DQN+Heuristic&218.08&77.0&228.52&0.86 \\
        \hline
        Clipped DQN&206.09&97.0&254.4&0.98 \\
        \hline
        Clipped DQN+Heuristic&259.7&98.0&248.33&0.9 \\
        \hline
    \end{tabular}
    \caption{Evaluation results for different reinforcement learning algorithms.}
    \label{tab:1}
\end{table*}

\subsection{Quantitative Analysis}
The evaluation results for all the algorithms are shown in Table \ref{tab:1}. 

We first start our evaluation on the three classic reinforcement learning algorithm with only one tiling, as shown in the first three rows of Table \ref{tab:1}. In this setting
$r_x = (0.5, 0.5, 0.5, 0.5, 0.2, 0.2, 0, 0)$. Both the initial learning rate and decay factor $\gamma$ are 0.9 for all 3 algorithms. The learning rate will decay for every 1000 episodes $5/(5+t)$ with the constant factor $c=5$. The maximum iteration of the training episodes is 5000. The final trained model at the end of the training is used to evaluate their performance. Among them, Q learning algorithm achieves the best result. However, their overall performance are still way poorer than the Deep learning approach. The Monte Carlo algorithm preform the worst. We discover that the performance of our algorithm is very sensitive to the discretization factor $r_x$. If the resolution of each variable is too high (i.e. value in $r_x$ is too small), then it is less likely for us to revisit a state and update their Q value. It will thus make the algorithm harder to converge. 

Note that the performance of MC is very poor. This is maybe caused by the high variance natural of the MC algorithm. We have tried to tune the learning rate, resolution, and the $\gamma$ parameters in the algorithm, but none of them can result in a good result. This MC is the every visit version. The performance of the first visit MC is worse than this. We suspect that this is because the number of samples for the MC is very little for it to find any meaningful result. Because the initial value of the Q function is zero, and zero action means do nothing. We think this is Q value with zero always dominant the action decision. this is why the MC decides to do nothing. 

Among all the algorithm we experimented, Heuristic DQN has the best performance with total $99\%$ probability of success and $258.23$ average score. This demonstrates the power of heuristic functions and shows that the proposed algorithm can help the network learn from prior knowledge during the early stage of training, leading to better performance. Also notice that all the three heuristic guided algorithms achieves satisfactory results. Heuristic DQN has improved near $60$ average scores and $30\%$ more accuracy. Also the performance of Clipped DQn is very good. Also we observe that even the worst NN-based methods outperform all the classical methods. And among the NN-based methods, Double DQN achives lowest average score which is $104.82$. We think this may be due to that double DQN uses two different networks and may requires more parameter fine tuning. This demonstrate the drawback of NN-based methods which lack generalization ability. However for the heuristic Double DQN, the average score is $228.52$, which is twice of the original double DQN's score, showing that heuristic can help network gain better generalization ability. For the average terminal score, the Heuristic DQN also achieves the best. Even though clipped DQN and heuristic DQN has similar average score and probability of success, the average fuel consumption for heuristic DQN is lower than clipped DQN. We believe it is because of the heuristic would help the robot navigate a more greedy way towards to the goal at the early stage of the training and this effect would help the network explore more direct way towards the goal with fewer fuels. 

We also study the impact from tile coding. We discover that when we starting using tile coding. We should make the value in $r_x$ larger, because the offset in tile coding will increase the resolution of the algorithm. In this experiment, we will use Q learning algorithm. We run three different resolutions with 2 tiles. 
\begin{enumerate}
    \item $res1 = (0.5, 0.5, 0.5, 0.5, 0.2, 0.2, 0, 0)$
    \item $res2 = (0.5, 0.5, 0.5, 0.5, 0.5, 0.5, 0, 0)$
    \item $res3 = (1.5, 1.5, 1.5, 1.5, 1.0, 1.0, 0, 0)$
\end{enumerate}
We find that the second resolution achieves the highest average score. Then with this second resolution, we run our algorithm with 4 tiles and 8 tiles. The resolution is shown in the Table \ref{tab:eval result for tile coding}. As we can see, when tile number equals to 4, we achieve the highest PoS among all the res2 cases. We also visualize the behavior of the this case, and we discover that the rocket will keep firing the main engine to maintain itself in the air, and gradually touch down. That is the reason why the average fuel consumption is so high. Note that it is not straightforward to compare this with our result in Table \ref{tab:1} because using the same resolution but with higher number of tile will increase the sparsity of our grid. It will make the algorithm harder to train and thus degrade the performance. 

\begin{table*}[t]
    \centering
    \begin{tabular}{|c|c|c|c|c|}
        \hline
        Coding Scheme&Average fuel consumption&Average terminal score&Average score&Probability of success\\
        \hline
        \hline
        T=2, res1&785.46&-11.04&-145.39&0.00 \\
        \hline
        T=2, res2&519.93&-53.98&-43.45&0.01 \\
        \hline
        T=2, res3&274.39&-91.11&-64.27&0.02 \\
        \hline
        T=4, res2&484.53&-31.98&-10.98&0.05 \\
        \hline
        T=8, res2&808.85&2.93&-54.05&0.00 \\
        \hline
    \end{tabular}
    \caption{Evaluation results for different tile coding.}
    \label{tab:eval result for tile coding}
\end{table*}

We study the value of the $\epsilon$ and how would it influence the behavior of the agent. The result is shown in Table \ref{tab:eval result for epsilon}. As we can see, both the $\epsilon=0.00$ and $\epsilon=0.05$ have high PoS. But in terms of the average score, the smaller the $\epsilon$, it better it is. This means that the smaller the $\epsilon$ is, the more stable the algorithm decision is.

\begin{table*}[t]
    \centering
    \begin{tabular}{|c|c|c|c|c|}
        \hline
        Coding Scheme&Average fuel consumption&Average terminal score&Average score&Probability of success\\
        \hline
        $\epsilon$=0.00&294.47&-31.96&43.30&0.15 \\
        \hline
        $\epsilon$=0.01&320.97&-26.90&19.64&0.07 \\
        \hline
        $\epsilon$=0.05&290.89&-41.98&-13.38&0.16 \\
        \hline
    \end{tabular}
    \caption{Evaluation results for different $\epsilon$.}
    \label{tab:eval result for epsilon}
\end{table*}

\subsection{NN-based methods}
\begin{figure}
    \centering
    \includegraphics[width=0.6\linewidth]{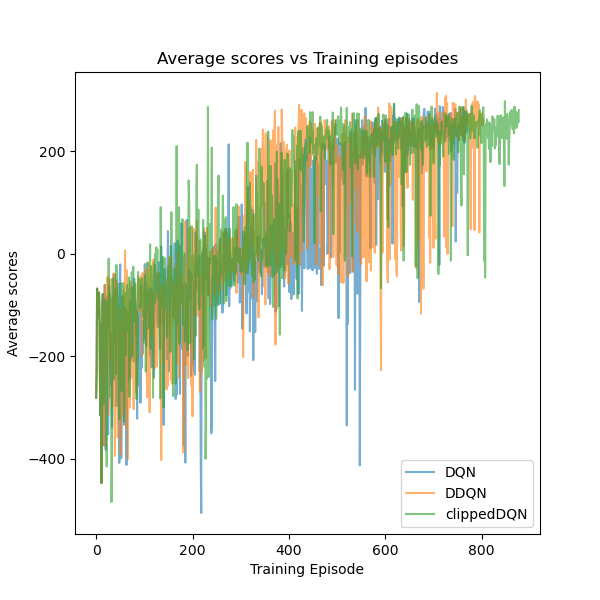}
    \caption{Average scores vs training episodes with batch size as 64. The figure is generated by plotting the average score for each episode.}
    \label{fig:1}
    \end{figure}

\begin{figure}
    \centering
    \includegraphics[width=0.6\linewidth]{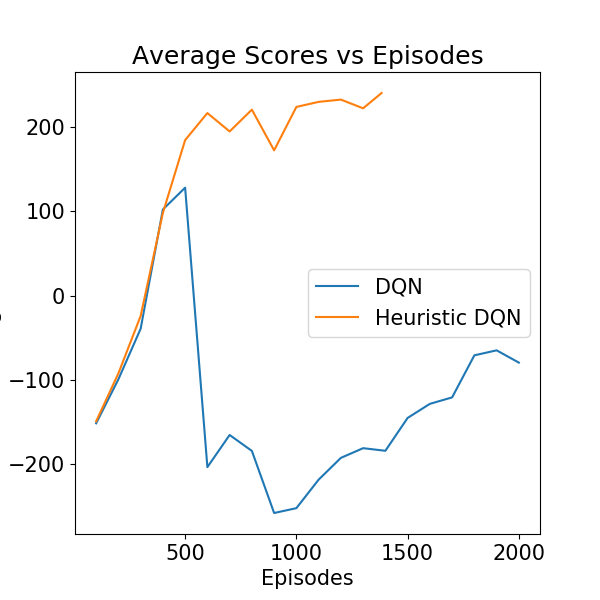}
    \caption{Average scores vs training episodes with batch size as 1024. The figure is generated by plotting the average score for every 100 episode.}
    \label{fig:2}
\end{figure}

The training curve for DQN, Double DQN and Clipped Double Q-Learning is shonw in the Figure \ref{fig:1}. The training average scores increases with large oscillation. This is because we don't have ground truth expected long term rewards. Unlike classical supervised learning tasks which have a ground truth value, for DQN based methods, they use bootstrapping, thus the network may oscillate as the target changes due to the update of its own parameters. And from Figure \ref{fig:1}, we can see that clipped DQN converges slightly faster with relatively low oscillation.  
To demonstrate the power of heuristic based methods, we conduct a comparison experiments. The hyper-parameters for DQN and heuristics DQN are same. And the batch size is 1024. As shown in Figure \ref{fig:2}, DQN will diverge while Heuristic DQN can still converge and successfully solve the problem. As we can see during early training stage, the curve of DQN increases but once DQN got lost due to the large number batches. It can't learn any useful information. However for Heuristic based methods, we can use heuristic function to guide the robot to navigate the search space. This prior knowledge helps the robot to move towards the goal when it does not have sufficiently konwledge about the problem. Once the robot have learnt enough useful information to guide it self towards goal, the decaying factor of heuristic function would gradually reduce the bias and the network can continue learning.

\subsection{Visualization}
\begin{enumerate}
    \item \textbf{Q-value}\\
    To better understand what neural networks have learned,  I choose to visualize the Q-value for Heuristic DQN. For each action, I plot a figure as shown in Figure \ref{fig:q-1}. I fixed all the other dimension in the state and only plot it vs x,y components for better visualization. 
        \begin{figure}[H]
        \centering
        \subfigure[action 0]{\includegraphics[width=0.45\linewidth]{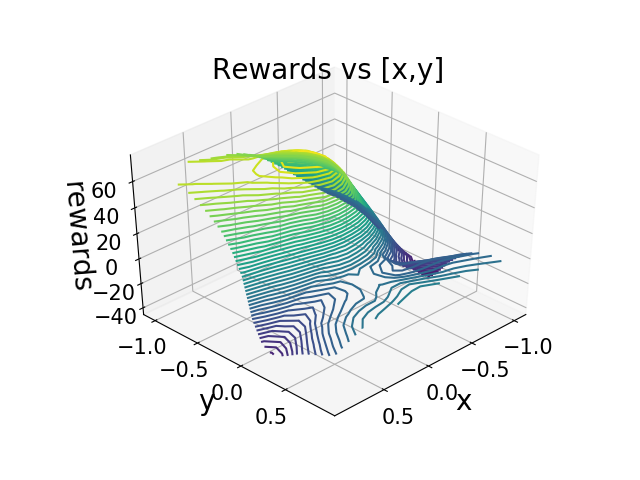}}
        \subfigure[action 1]{\includegraphics[width=0.45\linewidth]{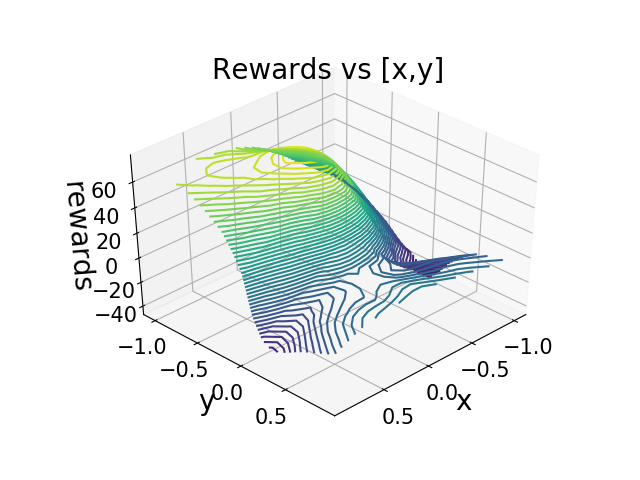}}
        \subfigure[action 2]{\includegraphics[width=0.45\linewidth]{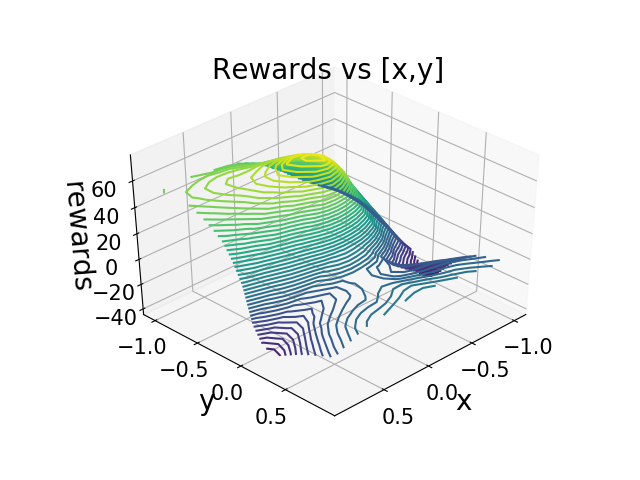}}
        \subfigure[action 3]{\includegraphics[width=0.45\linewidth]{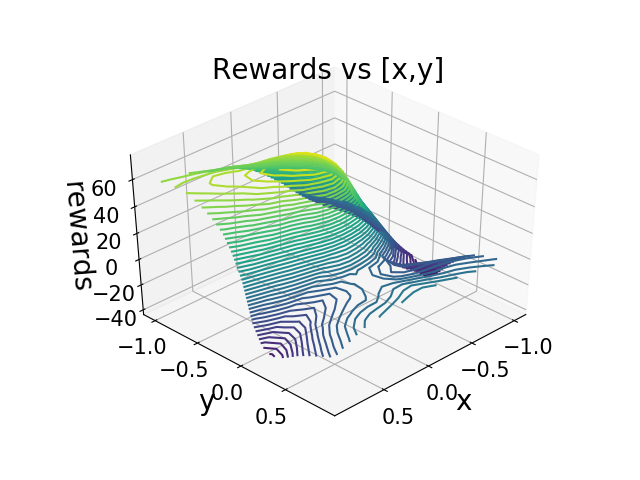}}
        \caption{Q-value visualization for Heuristic DQN when $state=[x,y,0,0,0,0,0,0]$. The height of z-axis corresponds to the value of Q. The highest point is near $[0,0]$. The range of $x,y$ is $-1$ to $1$ with step size $0.1$. }
        \label{fig:q-1}
        \end{figure}
        As we can see, the highest point is at $[0,0]$. And when $y$ decreases to 0, the Q-value increases. As $x$ moves towards $0$, the Q-value also increases. Since lunar lander always put landing pad at $[0,0]$, thus the highest q-value should around zero. Our experiment results agree with the fact, which means that our algorithm can learn correct q-value from a data-driven approach.
    \item \textbf{Game Visualization}\\
        We have generated \texttt{gif} visualization for various algorithms and attached them in the code part. Please refer to the \texttt{gif} folder or the \texttt{README} file for better visualizing the motion process. Here we also attach some images during the motion process for analysis purpose. As shown in Figure \ref{fig:game-1},\ref{fig:game-2},\ref{fig:game-3},\ref{fig:game-4},\ref{fig:game-5},\ref{fig:game-6} the robot will generally move towards the goal and try to reach the goal state. And we also observe that only Double DQN go out of the goal state. This also agree with the Table \ref{tab:1} that Double DQN has lowest average scores and probability of success. And for DQN, the robot will change its orientation a lot while the heuristic DQN moves directly to the goal, demonstrating better performance.
        \begin{figure}[H]
        \centering
        \subfigure[frame 0]{\includegraphics[width=0.45\linewidth]{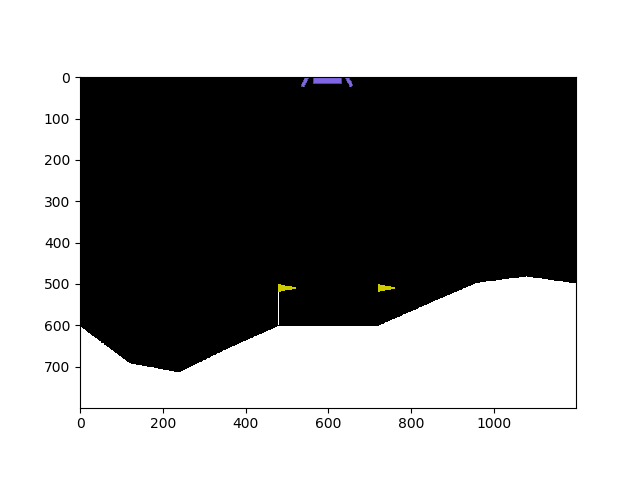}}
        \subfigure[frame 40]{\includegraphics[width=0.45\linewidth]{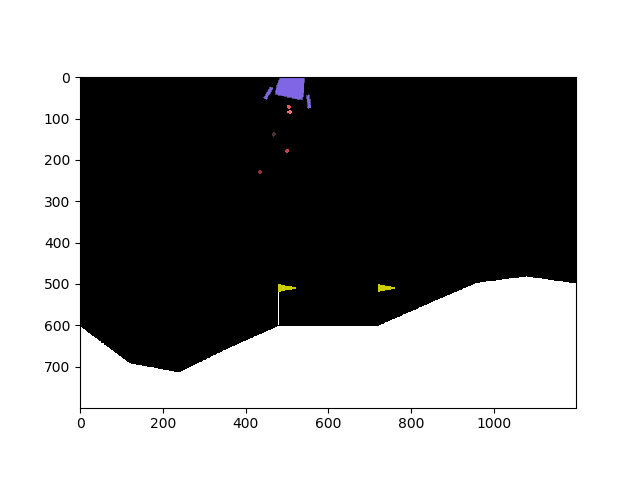}}
        \subfigure[frame 80]{\includegraphics[width=0.45\linewidth]{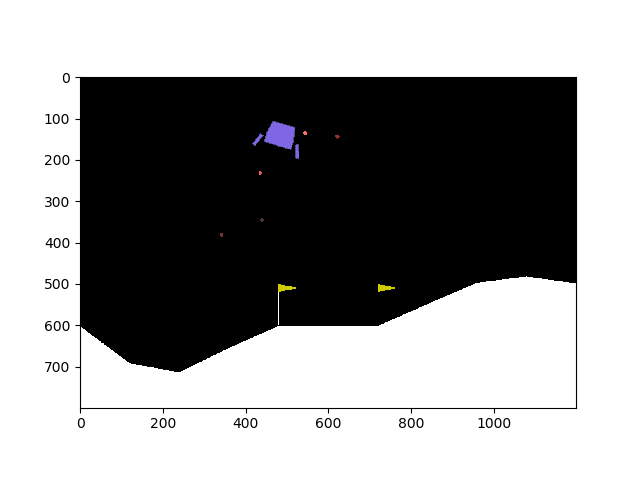}}
        \subfigure[frame 120]{\includegraphics[width=0.45\linewidth]{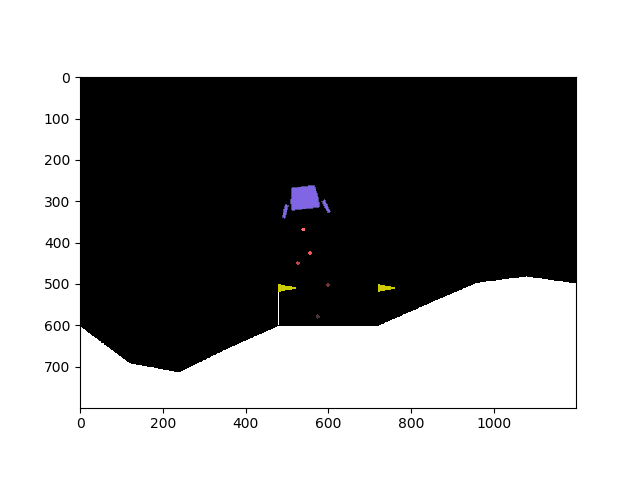}}
        \subfigure[frame 160]{\includegraphics[width=0.45\linewidth]{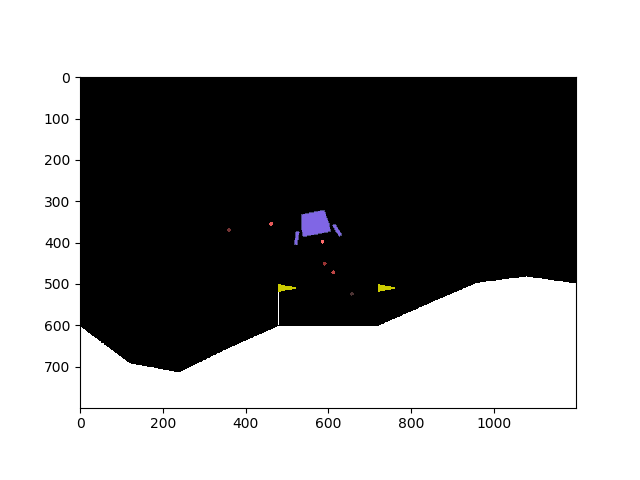}}
        \subfigure[frame 200]{\includegraphics[width=0.45\linewidth]{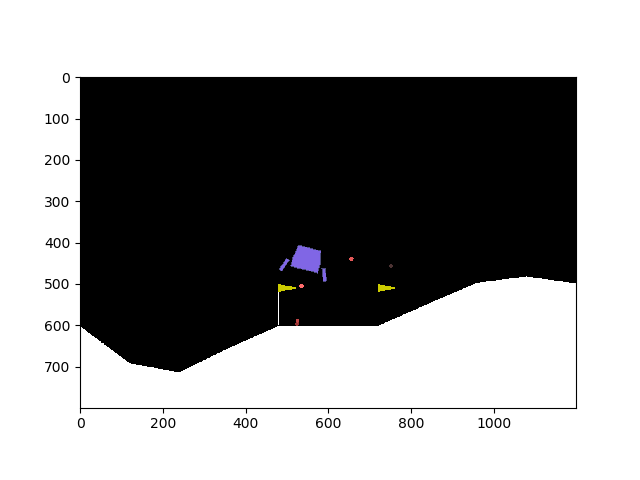}}
        \subfigure[frame 240]{\includegraphics[width=0.45\linewidth]{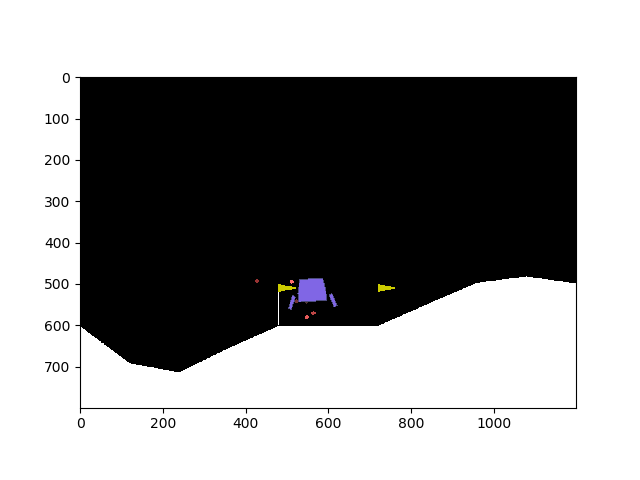}}
        \subfigure[frame 280]{\includegraphics[width=0.45\linewidth]{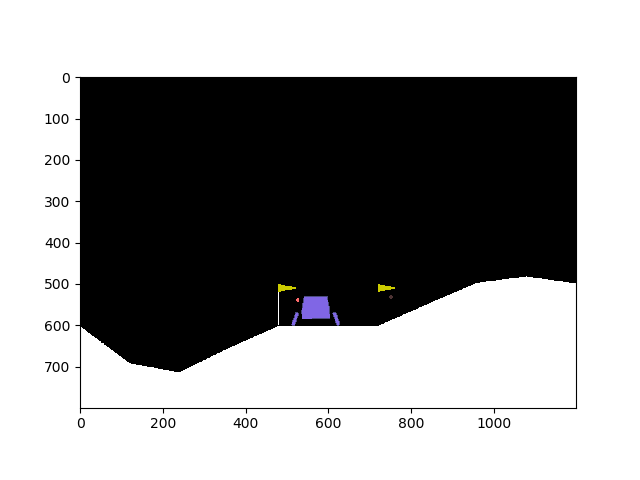}}
        \caption{Game visualization for DQN. Please refer to the attached gif file for complete motion process.}
        \label{fig:game-1}
        \end{figure}
        
        \begin{figure}[H]
        \centering
        \subfigure[frame 0]{\includegraphics[width=0.45\linewidth]{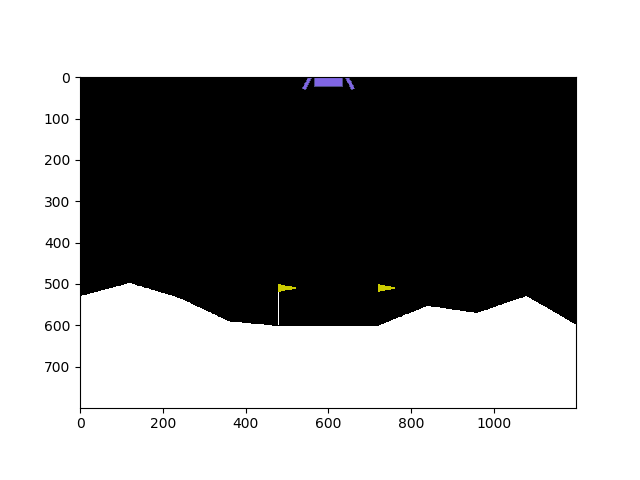}}
        \subfigure[frame 40]{\includegraphics[width=0.45\linewidth]{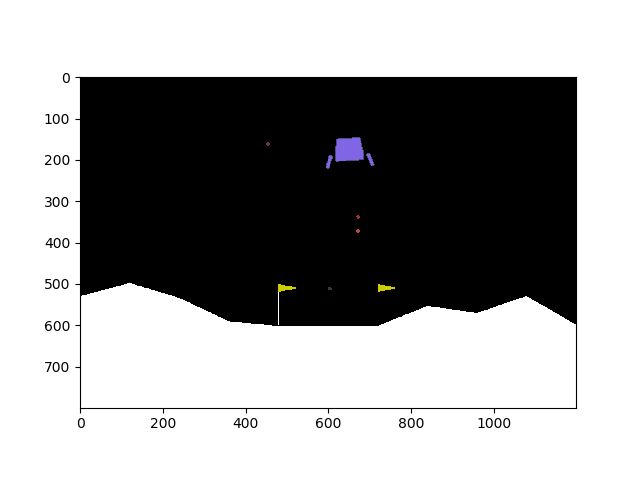}}
        \subfigure[frame 80]{\includegraphics[width=0.45\linewidth]{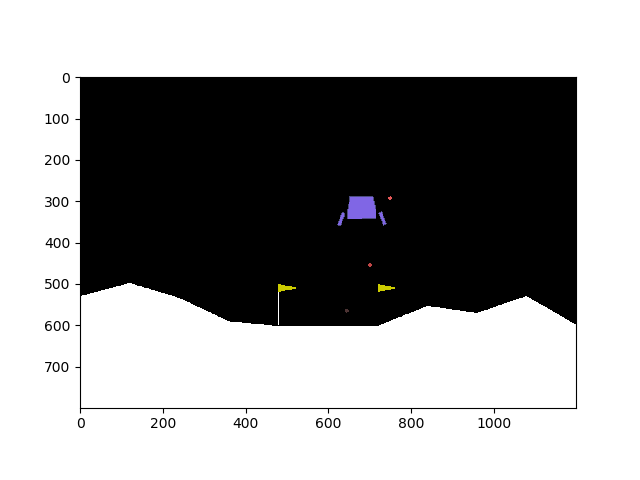}}
        \subfigure[frame 120]{\includegraphics[width=0.45\linewidth]{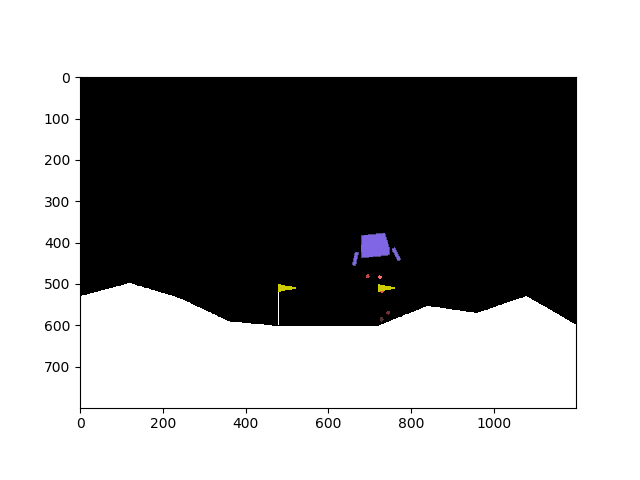}}
        \subfigure[frame 160]{\includegraphics[width=0.45\linewidth]{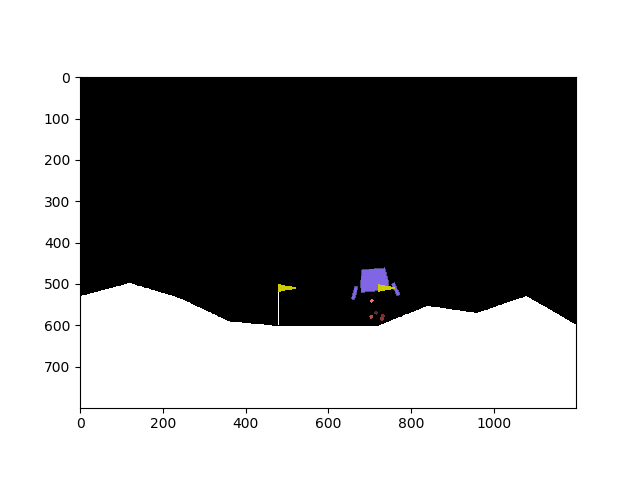}}
        \subfigure[frame 200]{\includegraphics[width=0.45\linewidth]{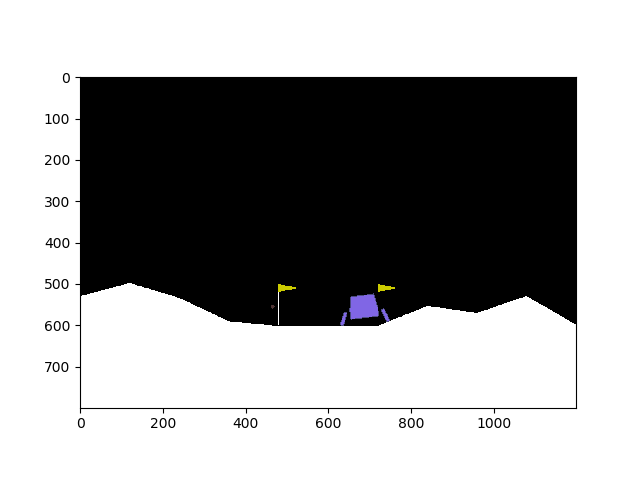}}
        \subfigure[frame 240]{\includegraphics[width=0.45\linewidth]{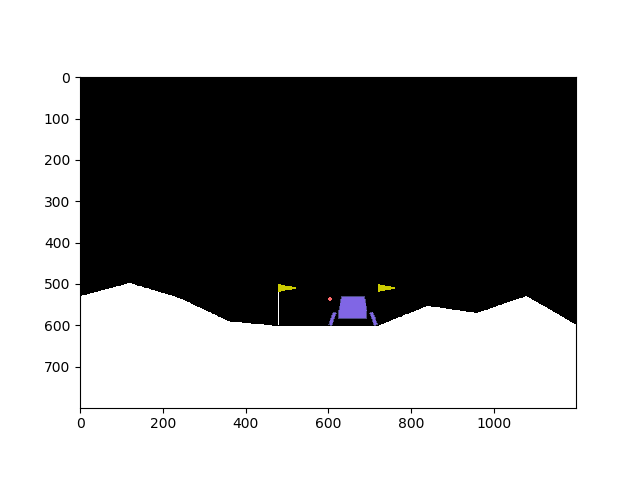}}
        \subfigure[frame 280]{\includegraphics[width=0.45\linewidth]{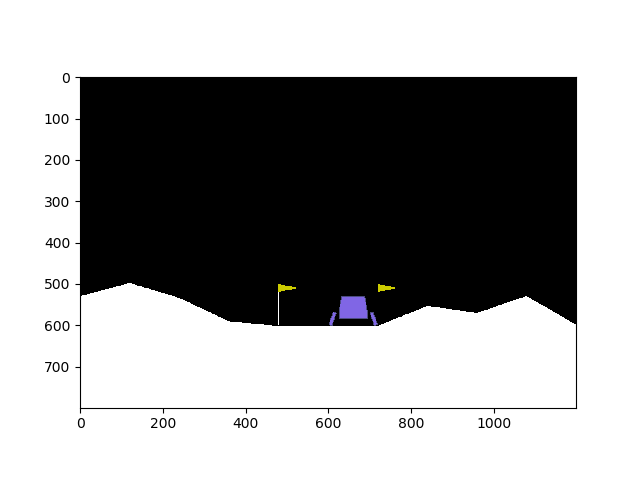}}
        \caption{Game visualization for Heuristic DQN. }
        \label{fig:game-2}
        \end{figure}
        
        \begin{figure}[H]
        \centering
        \subfigure[frame 0]{\includegraphics[width=0.45\linewidth]{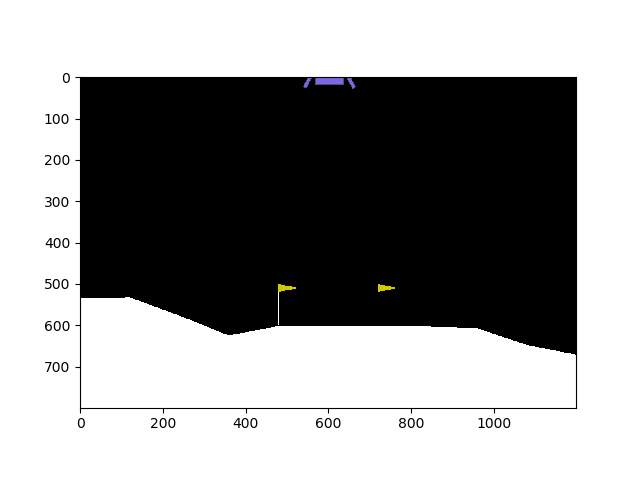}}
        \subfigure[frame 20]{\includegraphics[width=0.45\linewidth]{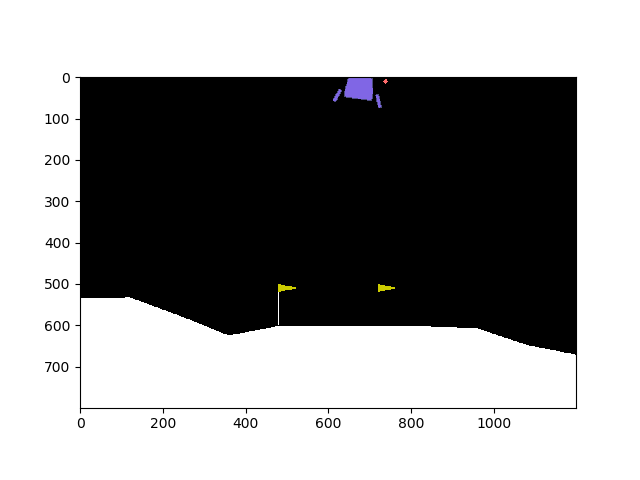}}
        \subfigure[frame 40]{\includegraphics[width=0.45\linewidth]{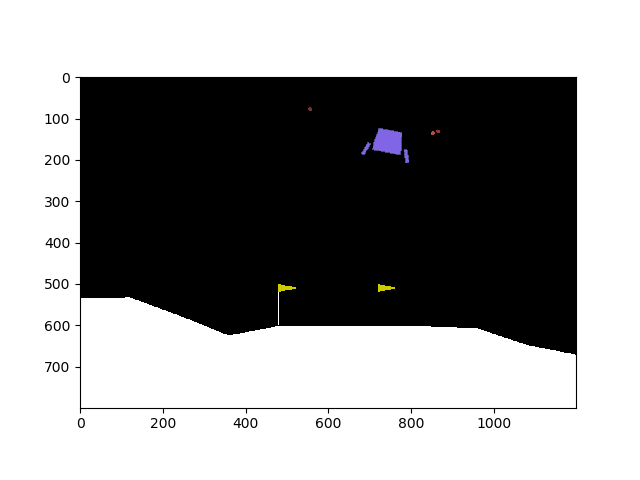}}
        \subfigure[frame 60]{\includegraphics[width=0.45\linewidth]{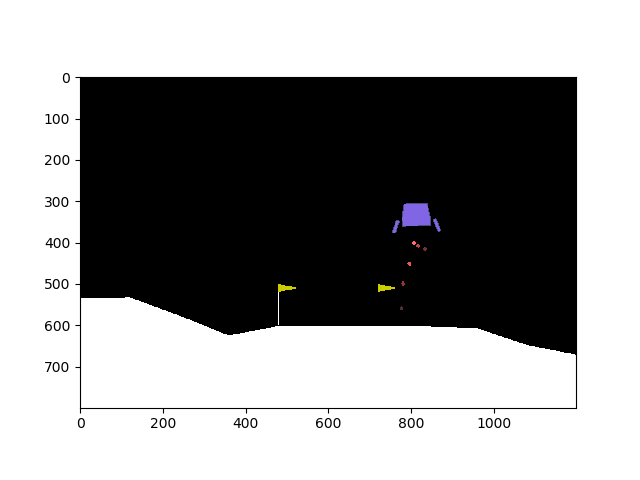}}
        \subfigure[frame 80]{\includegraphics[width=0.45\linewidth]{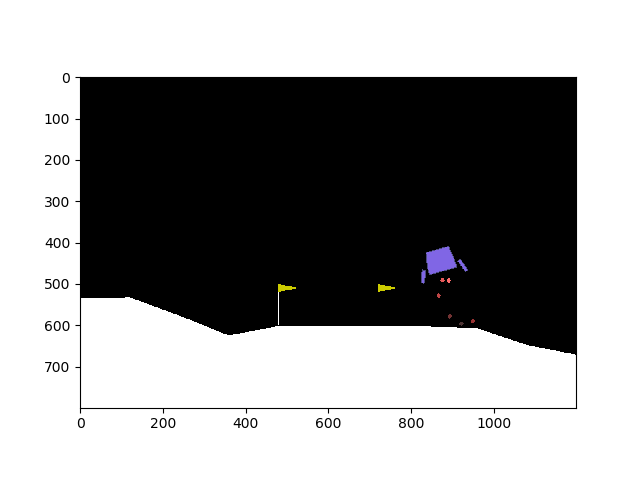}}
        \subfigure[frame 100]{\includegraphics[width=0.45\linewidth]{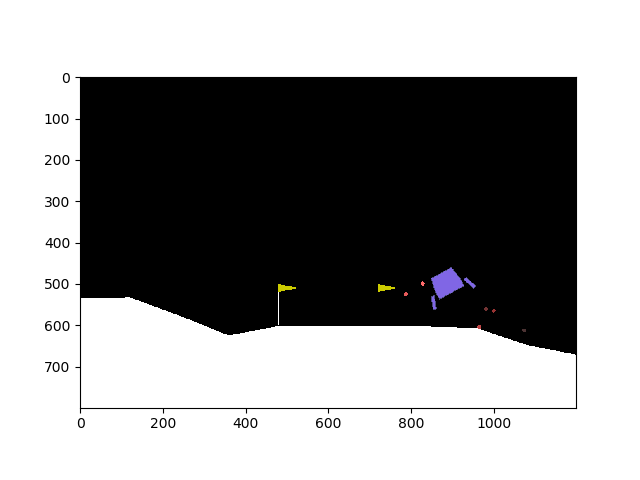}}
        \subfigure[frame 120]{\includegraphics[width=0.45\linewidth]{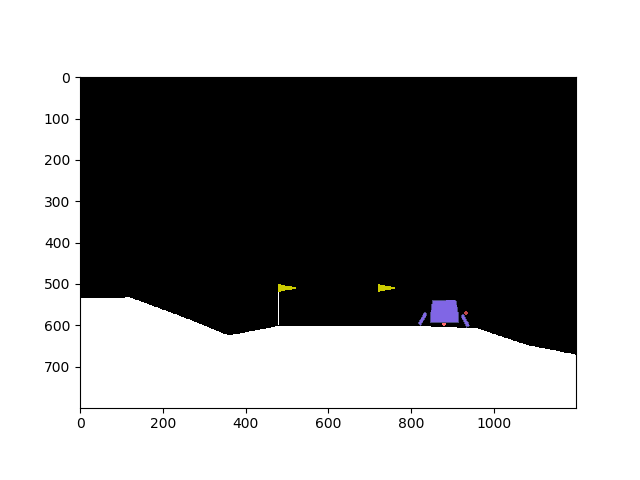}}
        \caption{Game visualization for Double DQN. }
        \label{fig:game-3}
        \end{figure}
        
        \begin{figure}[H]
        \centering
        \subfigure[frame 0]{\includegraphics[width=0.45\linewidth]{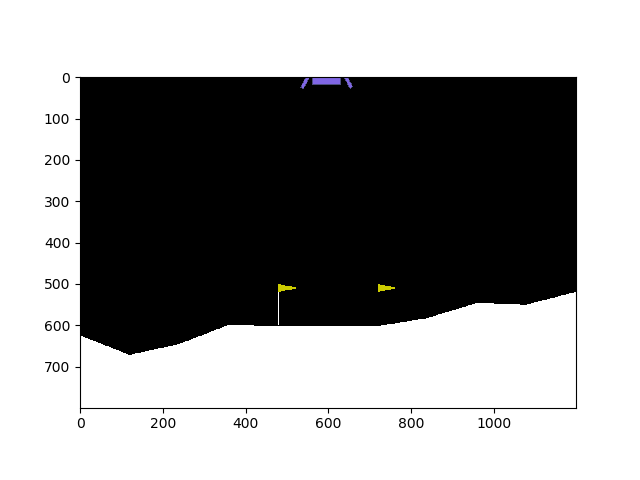}}
        \subfigure[frame 40]{\includegraphics[width=0.45\linewidth]{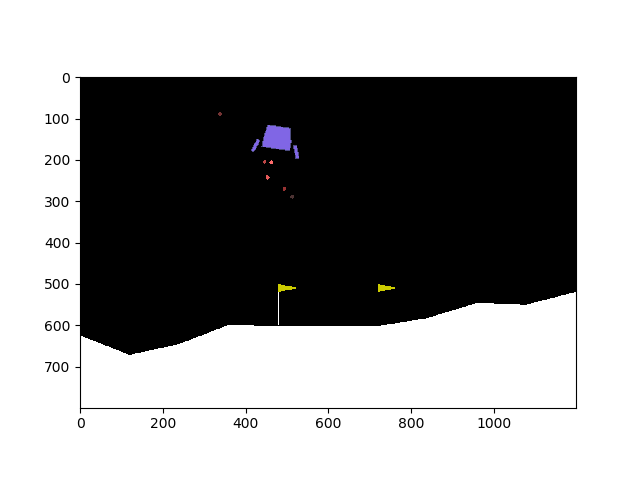}}
        \subfigure[frame 80]{\includegraphics[width=0.45\linewidth]{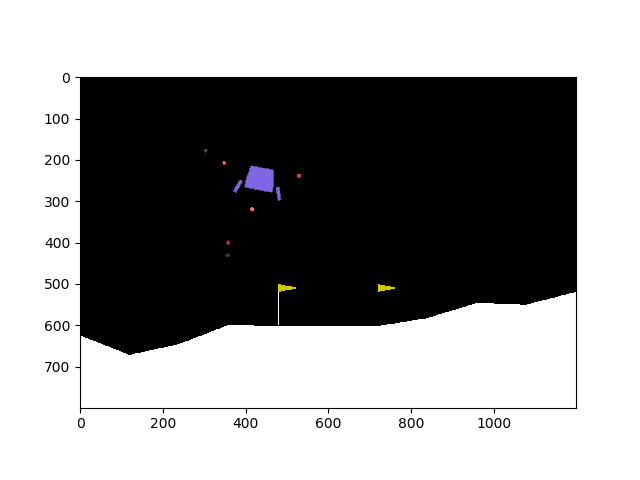}}
        \subfigure[frame 120]{\includegraphics[width=0.45\linewidth]{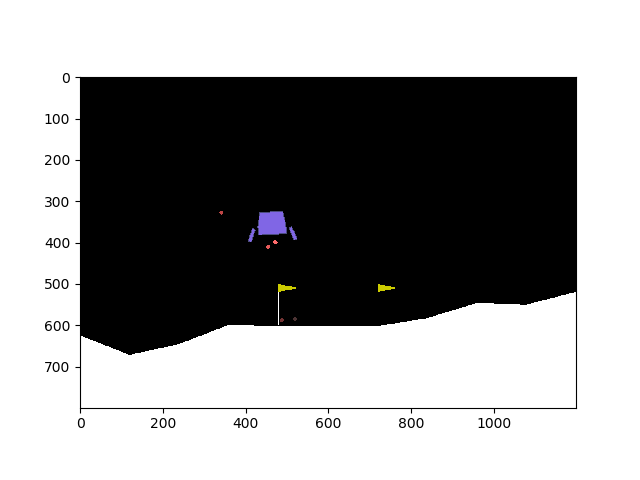}}
        \subfigure[frame 160]{\includegraphics[width=0.45\linewidth]{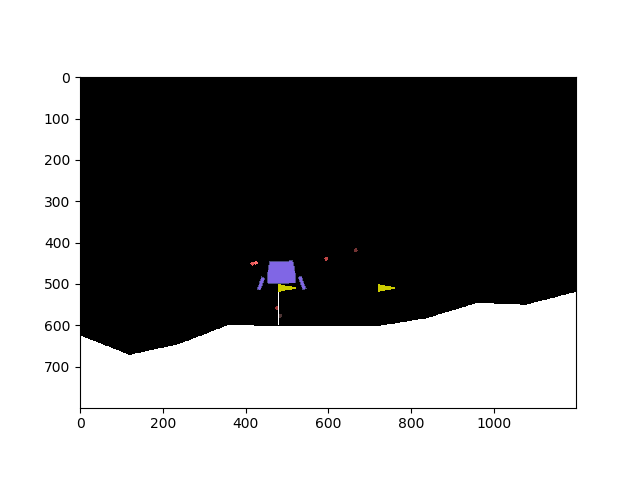}}
        \subfigure[frame 200]{\includegraphics[width=0.45\linewidth]{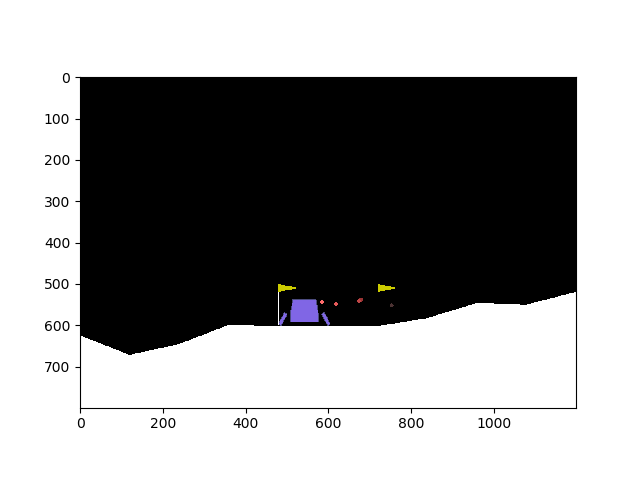}}
        \subfigure[frame 240]{\includegraphics[width=0.45\linewidth]{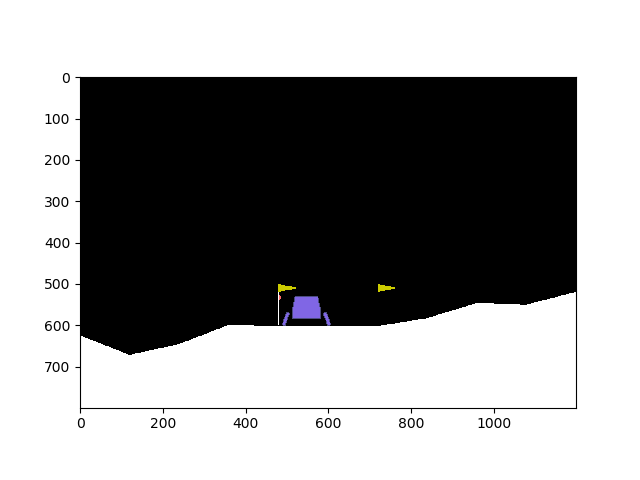}}
        \subfigure[frame 280]{\includegraphics[width=0.45\linewidth]{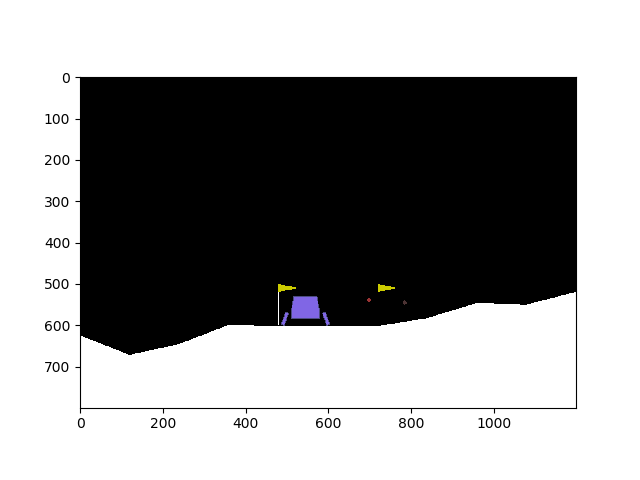}}
        \caption{Game visualization for Heuristic Double DQN. }
        \label{fig:game-4}
        \end{figure}
        
         \begin{figure}[H]
        \centering
        \subfigure[frame 0]{\includegraphics[width=0.45\linewidth]{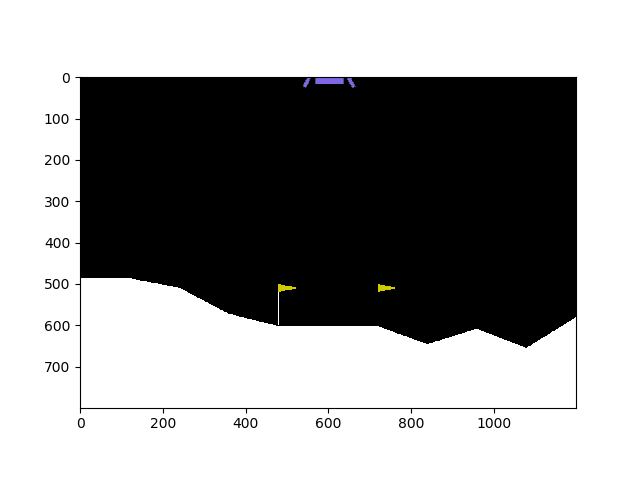}}
        \subfigure[frame 40]{\includegraphics[width=0.45\linewidth]{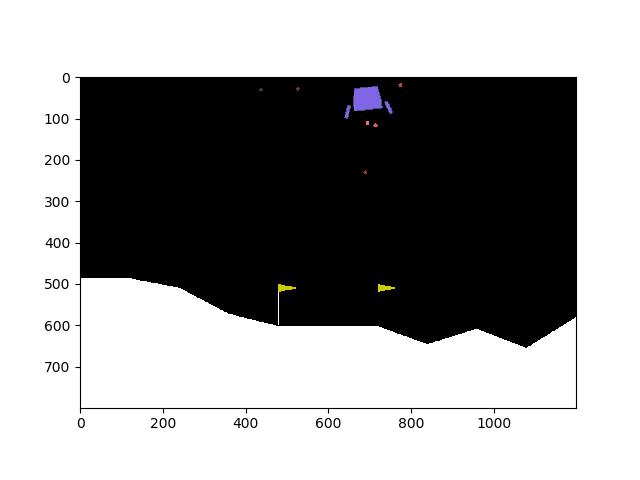}}
        \subfigure[frame 80]{\includegraphics[width=0.45\linewidth]{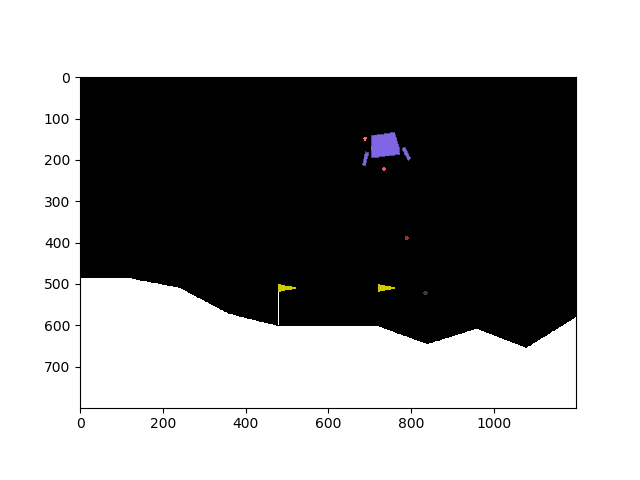}}
        \subfigure[frame 120]{\includegraphics[width=0.45\linewidth]{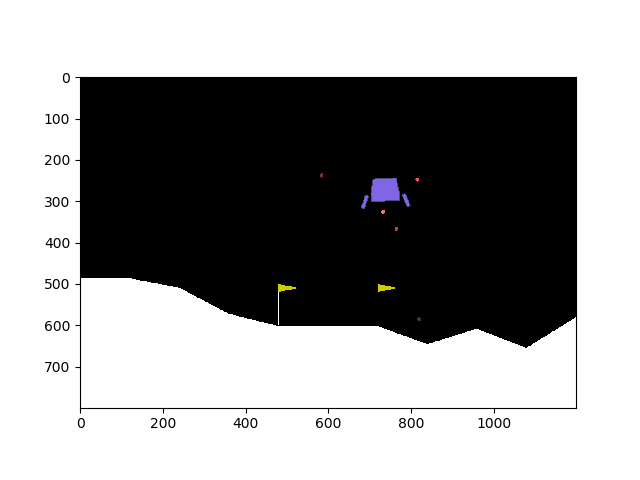}}
        \subfigure[frame 160]{\includegraphics[width=0.45\linewidth]{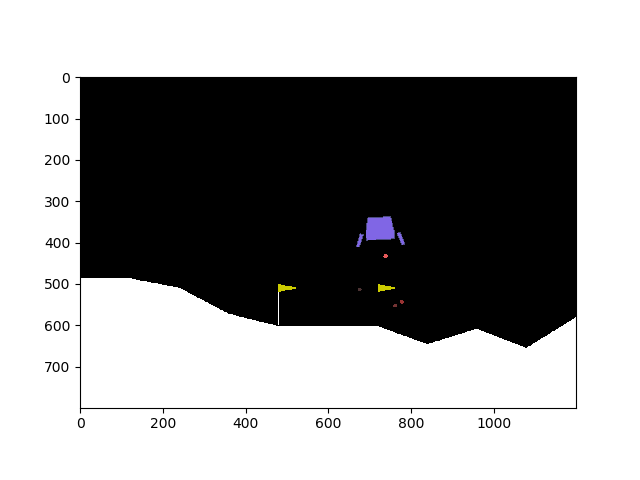}}
        \subfigure[frame 200]{\includegraphics[width=0.45\linewidth]{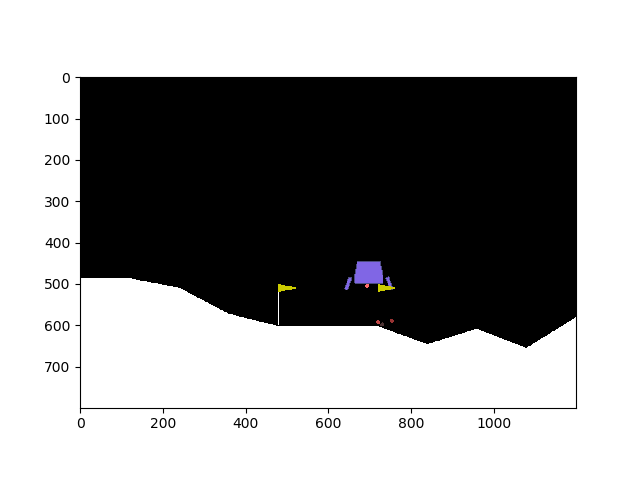}}
        \subfigure[frame 240]{\includegraphics[width=0.45\linewidth]{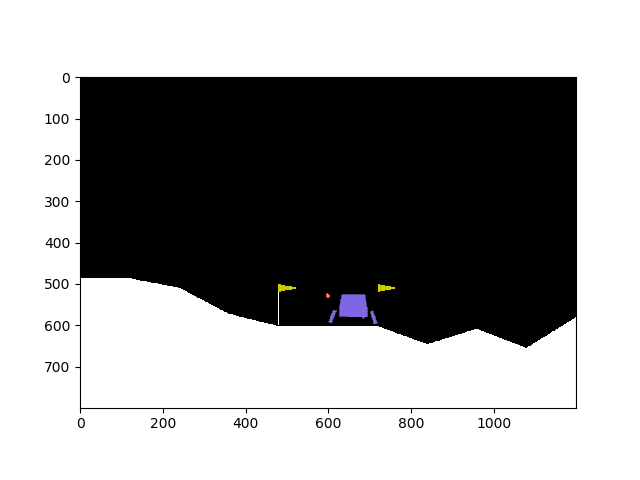}}
        \subfigure[frame 280]{\includegraphics[width=0.45\linewidth]{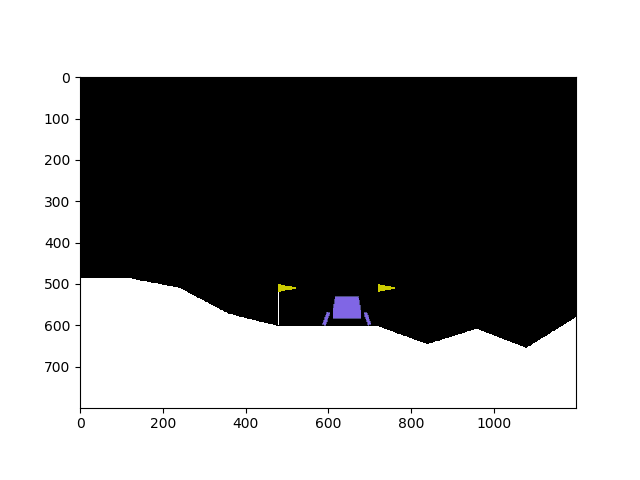}}
        \subfigure[frame 320]{\includegraphics[width=0.45\linewidth]{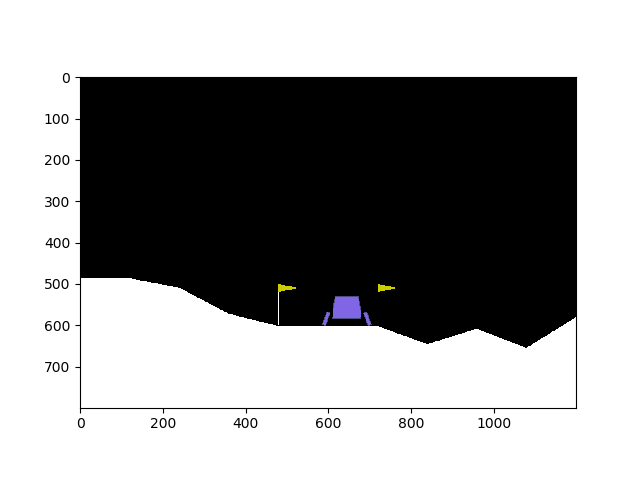}}
        \caption{Game visualization for Clipped Double Q-Learning. }
        \label{fig:game-5}
        \end{figure}
        
        \begin{figure}[H]
        \centering
        \subfigure[frame 0]{\includegraphics[width=0.45\linewidth]{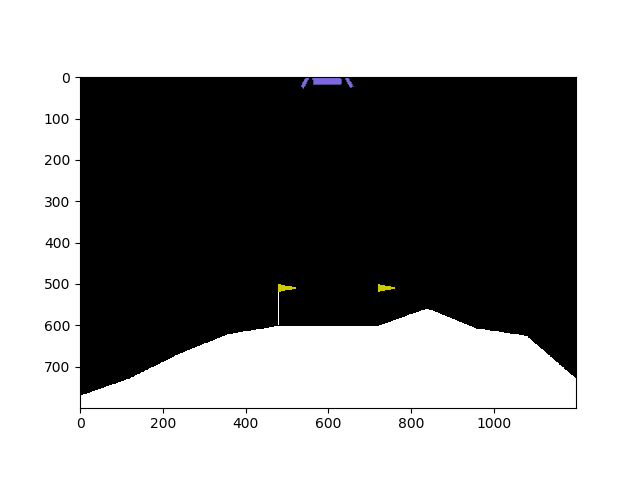}}
        \subfigure[frame 40]{\includegraphics[width=0.45\linewidth]{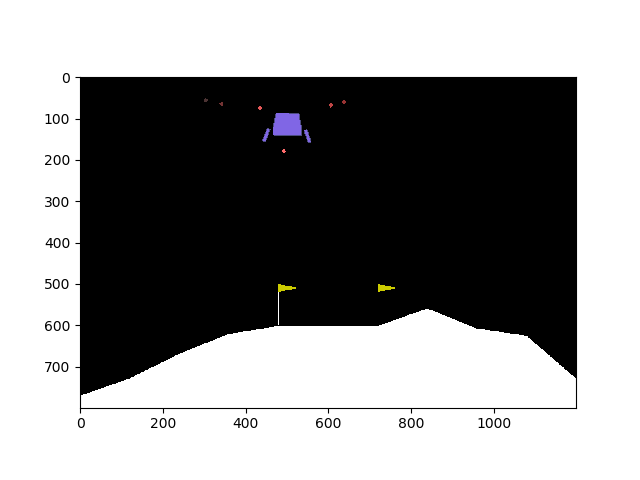}}
        \subfigure[frame 80]{\includegraphics[width=0.45\linewidth]{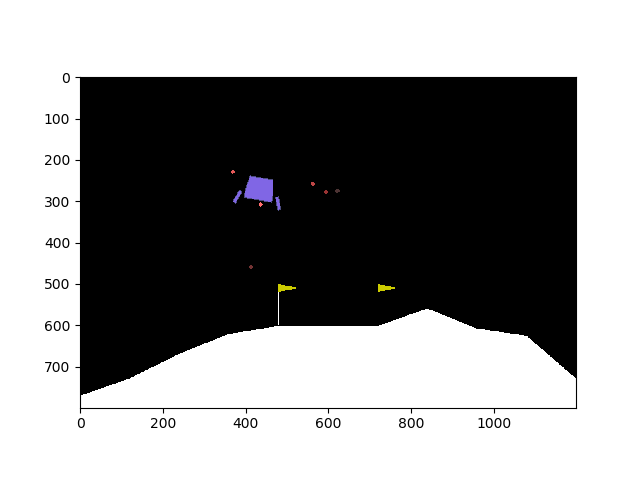}}
        \subfigure[frame 120]{\includegraphics[width=0.45\linewidth]{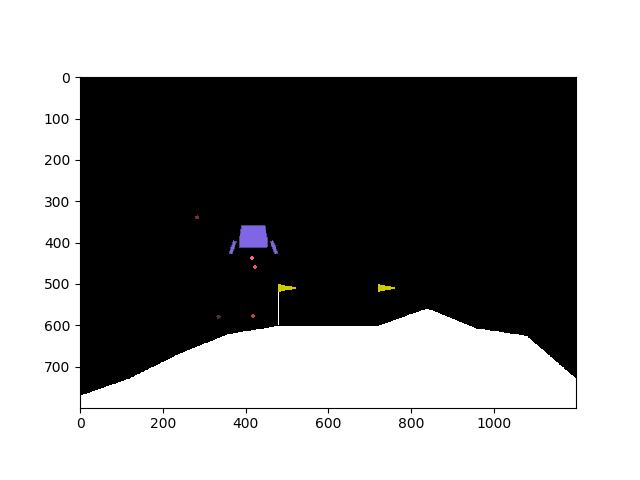}}
        \subfigure[frame 160]{\includegraphics[width=0.45\linewidth]{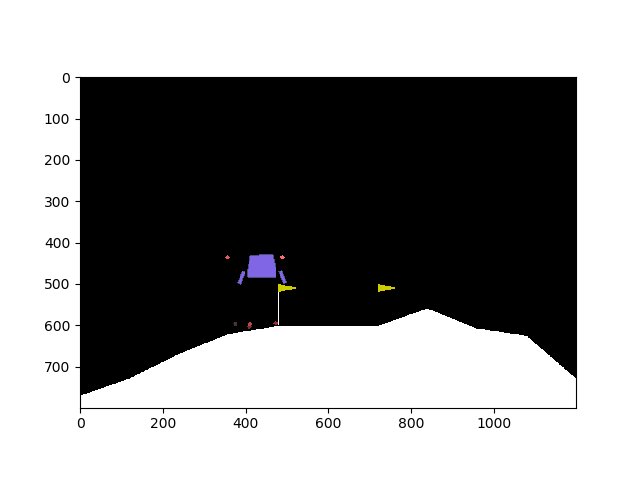}}
        \subfigure[frame 200]{\includegraphics[width=0.45\linewidth]{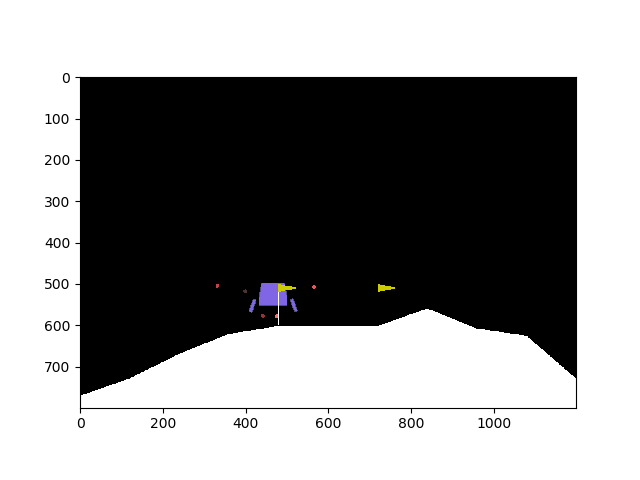}}
        \subfigure[frame 240]{\includegraphics[width=0.45\linewidth]{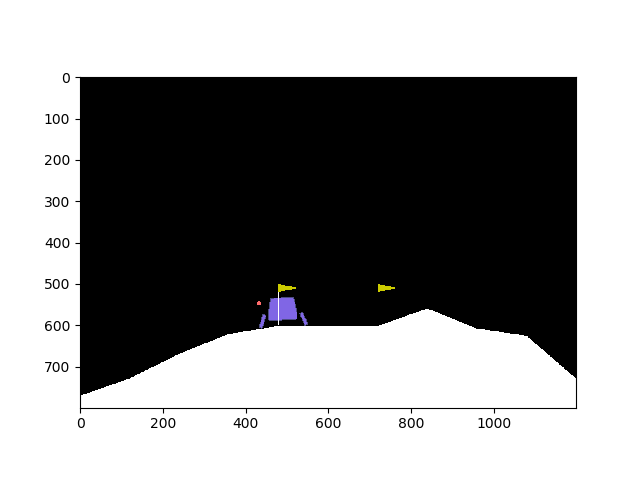}}
        \subfigure[frame 280]{\includegraphics[width=0.45\linewidth]{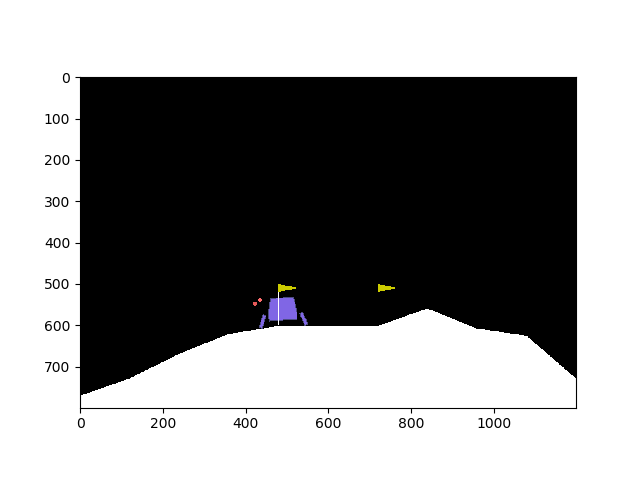}}
        \subfigure[frame 320]{\includegraphics[width=0.45\linewidth]{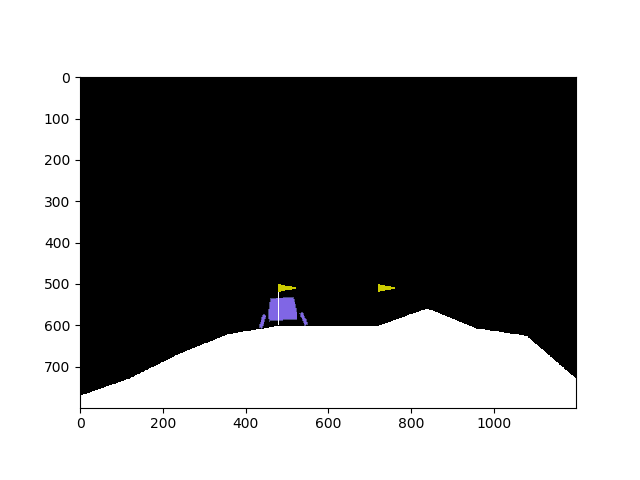}}
        \caption{Game visualization for Heuristic Clipped Double Q-Learning. }
        \label{fig:game-6}
        \end{figure}

        \begin{figure}[H]
        \centering
        \subfigure[frame 0]{\includegraphics[width=0.45\linewidth]{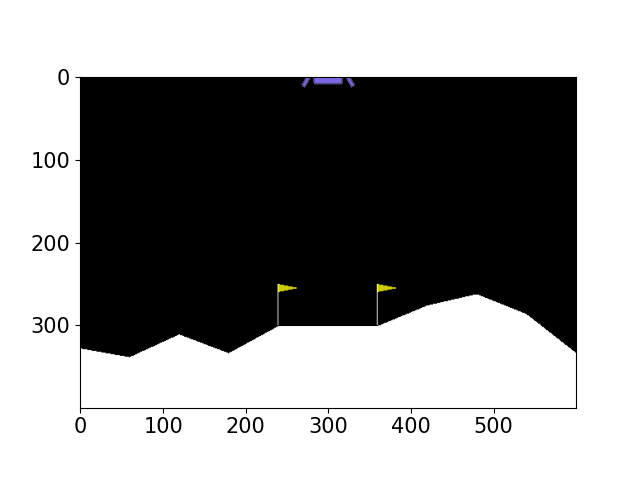}}
        \subfigure[frame 40]{\includegraphics[width=0.45\linewidth]{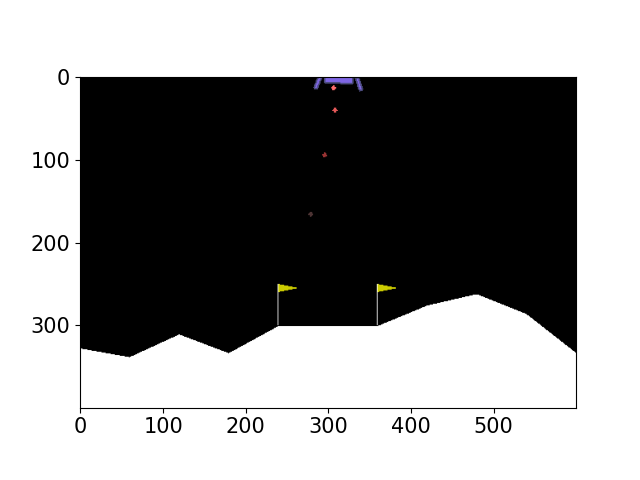}}
        \subfigure[frame 80]{\includegraphics[width=0.45\linewidth]{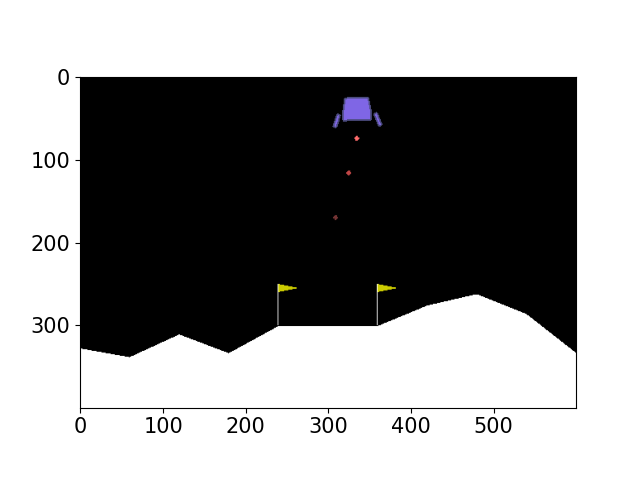}}
        \subfigure[frame 120]{\includegraphics[width=0.45\linewidth]{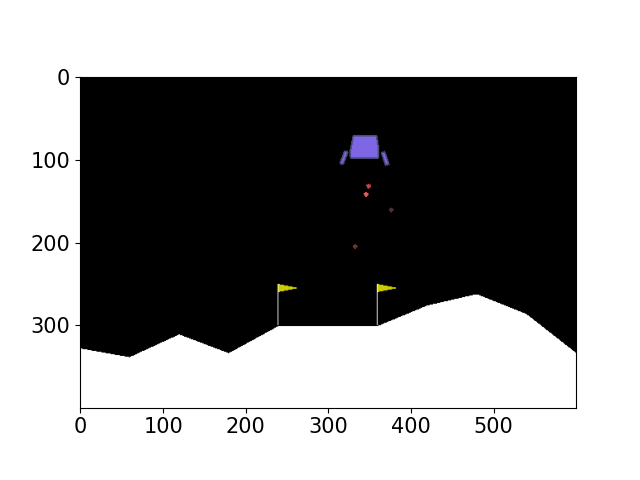}}
        \subfigure[frame 160]{\includegraphics[width=0.45\linewidth]{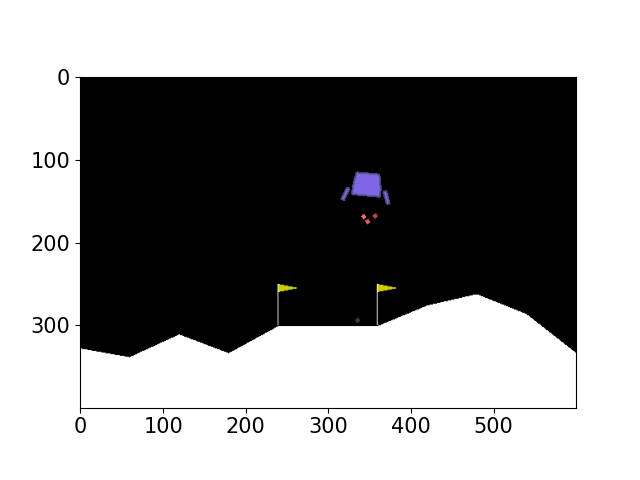}}
        \subfigure[frame 200]{\includegraphics[width=0.45\linewidth]{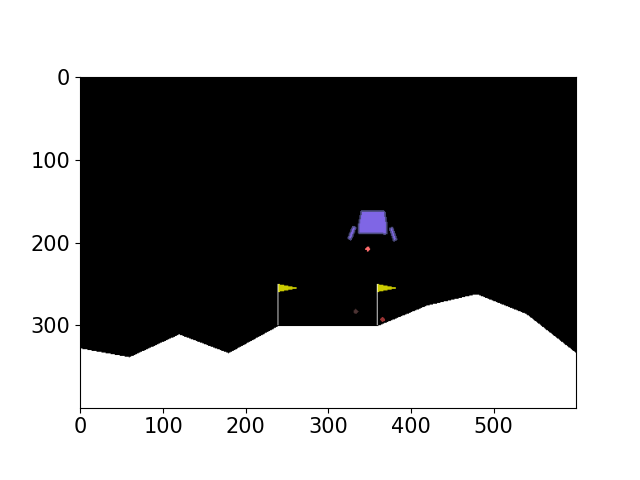}}
        \subfigure[frame 240]{\includegraphics[width=0.45\linewidth]{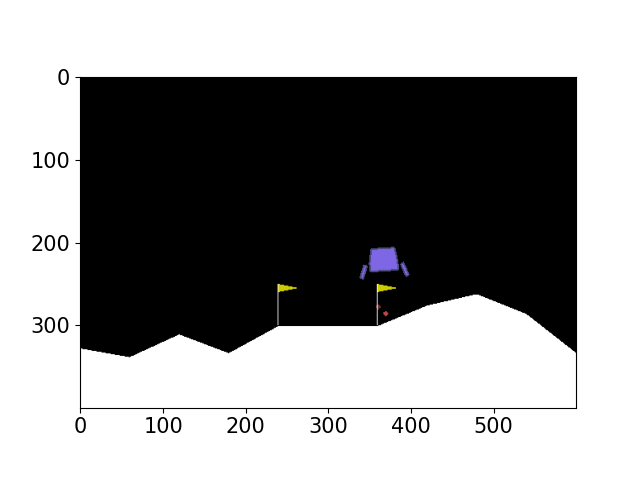}}
        \subfigure[frame 280]{\includegraphics[width=0.45\linewidth]{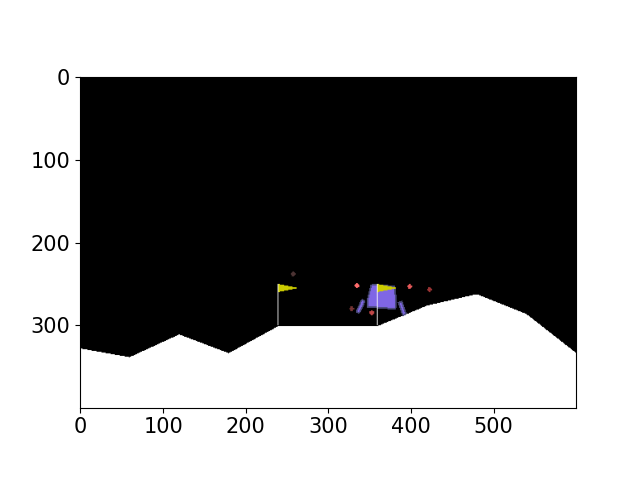}}
        \subfigure[frame 320]{\includegraphics[width=0.45\linewidth]{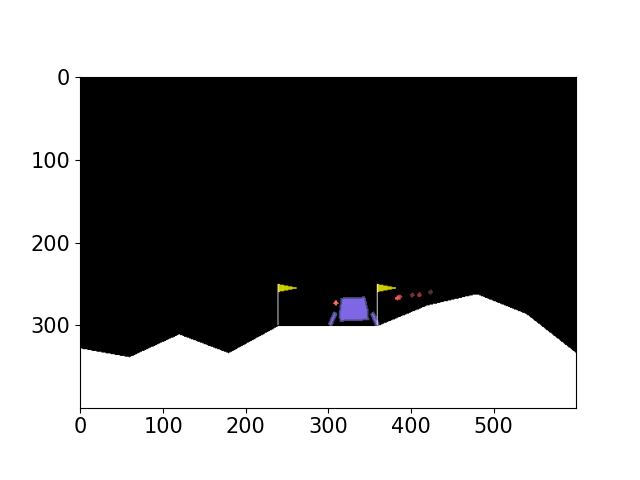}}
        \caption{Game visualization for Q-Learning.}
        \label{fig:QLearning}
        \end{figure}
        
        \begin{figure}[H]
        \centering
        \subfigure[frame 0]{\includegraphics[width=0.45\linewidth]{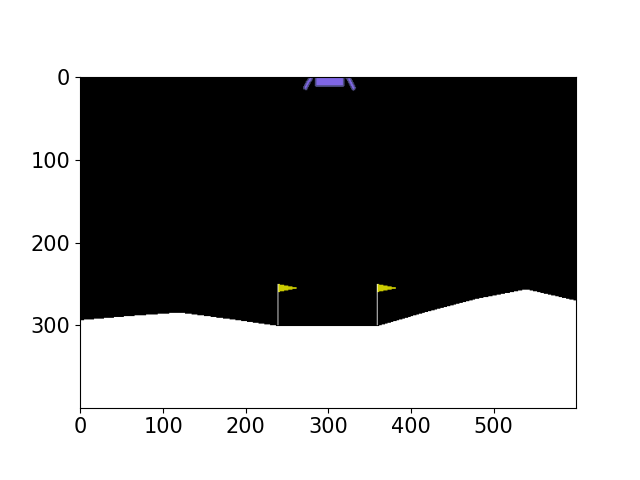}}
        \subfigure[frame 40]{\includegraphics[width=0.45\linewidth]{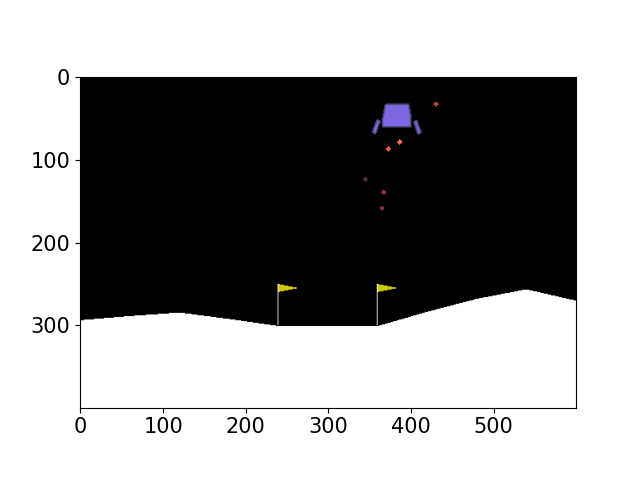}}
        \subfigure[frame 80]{\includegraphics[width=0.45\linewidth]{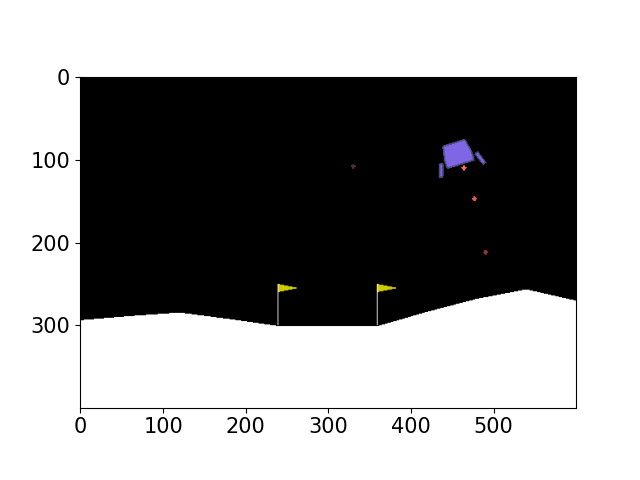}}
        \subfigure[frame 120]{\includegraphics[width=0.45\linewidth]{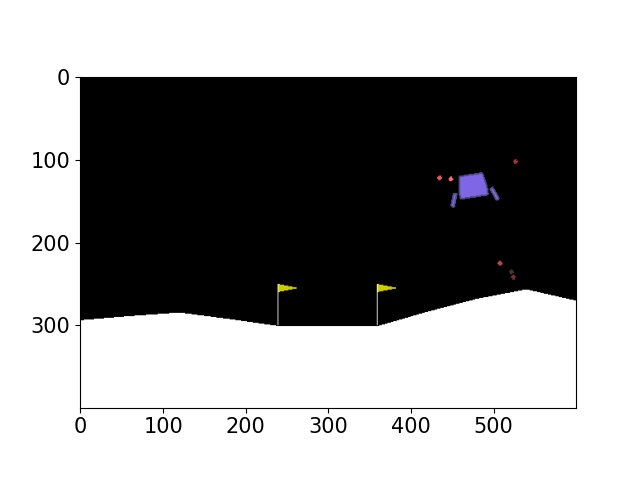}}
        \subfigure[frame 160]{\includegraphics[width=0.45\linewidth]{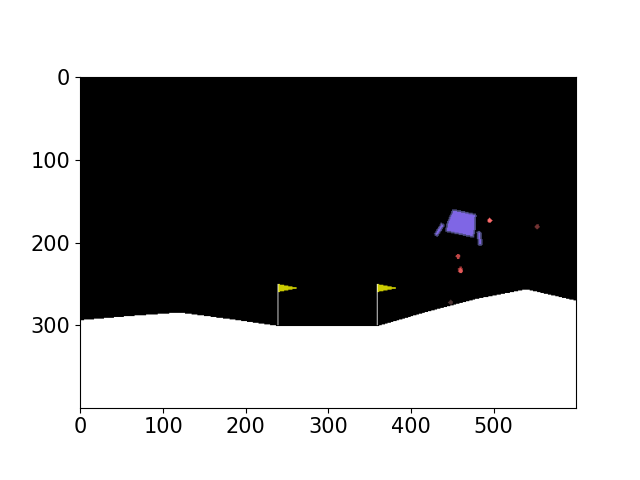}}
        \subfigure[frame 200]{\includegraphics[width=0.45\linewidth]{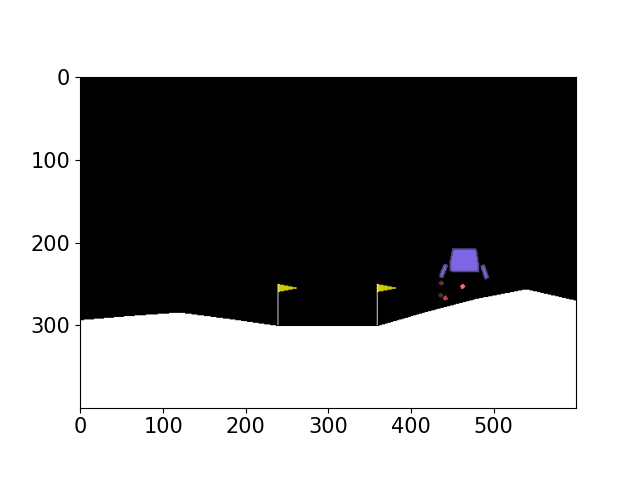}}
        \subfigure[frame 240]{\includegraphics[width=0.45\linewidth]{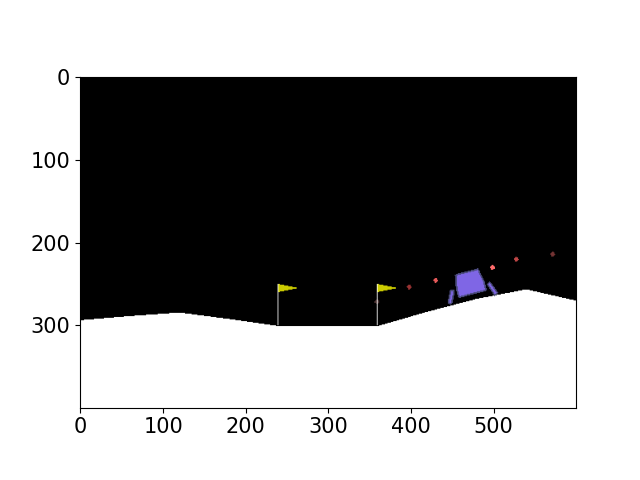}}
        \subfigure[frame 280]{\includegraphics[width=0.45\linewidth]{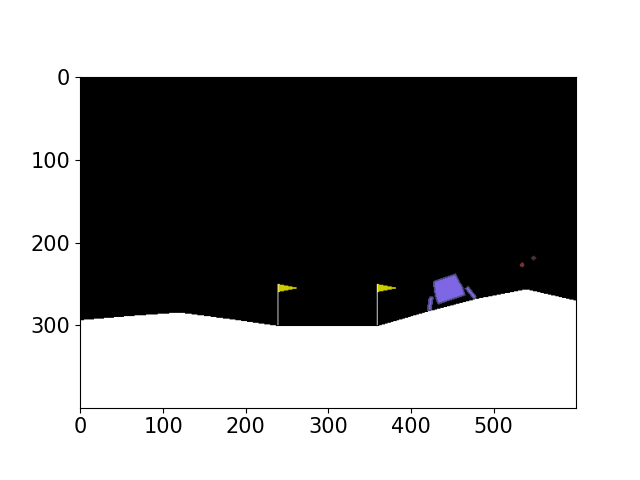}}
        \subfigure[frame 320]{\includegraphics[width=0.45\linewidth]{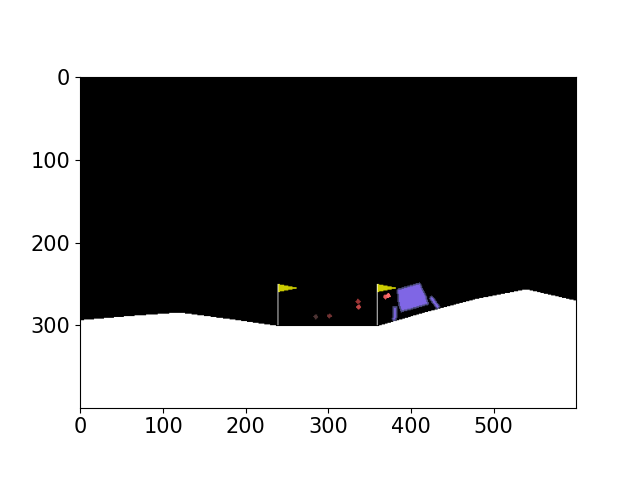}}
        \subfigure[frame 360]{\includegraphics[width=0.45\linewidth]{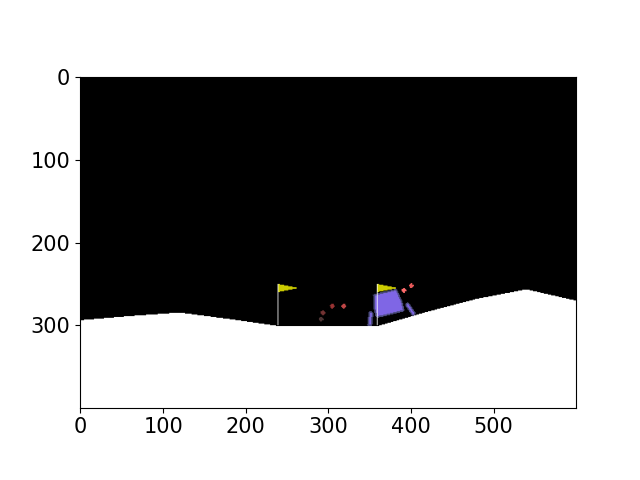}}
        \caption{Game visualization for SARSA.}
        \label{fig:SARSA}
        \end{figure}

        \begin{figure}[H]
        \centering
        \subfigure[frame 0]{\includegraphics[width=0.45\linewidth]{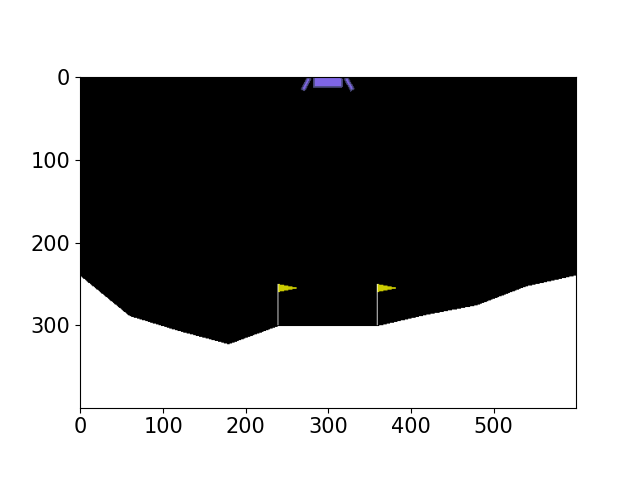}}
        \subfigure[frame 40]{\includegraphics[width=0.45\linewidth]{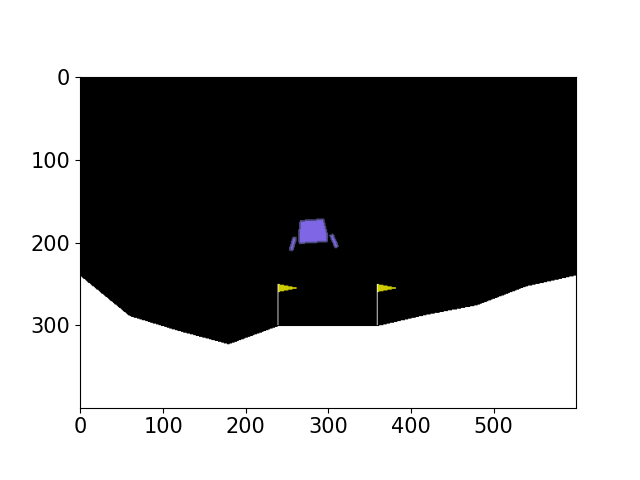}}
        \subfigure[frame 70]{\includegraphics[width=0.45\linewidth]{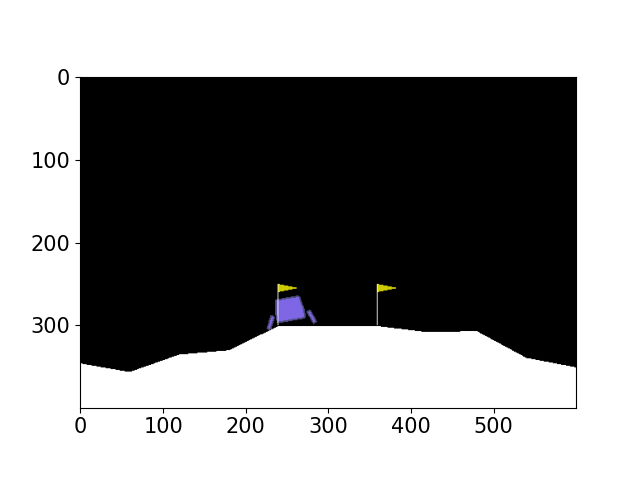}}
        \caption{Game visualization for MonteCarlo.}
        \label{fig:MonteCarlo}
        \end{figure}
    
\end{enumerate}
\section{Conclusion}
In this paper, we experimented many algorithms and proposed our own Heuristic Reinforcement Learning algorithm. And we have demonstrated the success of heuristic for many neural networks algorithms. Notice that our proposed algorithm can be easily transformed to classical methods. Also the introduced heuristic should play more roles in the early stage training to void the potential bad effect of human bias. And for latter training, we use decay factor to let algorithm rely more on its experience/learned features/bootstrapping. 
\bibliography{mybib}{}
\bibliographystyle{plain}
\end{document}